\documentclass[a4paper,fleqn]{cas-dc}

\usepackage{soul} 
\usepackage{color, xcolor}
\usepackage[numbers]{natbib}
\usepackage{subfigure}
\usepackage{tabularx}
\usepackage{tikz}
\usepackage{pgfplots}
\usepackage[edges]{forest}
\usepackage{adjustbox}
\usepackage{url}
\usepackage{graphicx}
\usepackage{array}
\usepackage{tabularx}

\begin{document}
\begin{sloppypar}
\shorttitle{Mixture of Experts (MoE): A Big Data Perspective} 
\shortauthors{W. Gan \textit{et al.}}

\author[1]{Wensheng Gan}
\ead{wsgan001@gmail.com}
\address[1]{College of Cyber Security, Jinan University, Guangzhou 510632, China}

\author[2]{Zhenyao Ning}
\ead{zyning5@gmail.com}
\address[2]{College of Mathematics and Computer Science, Shantou University, Shantou 515899, China}

\author[3]{Zhenlian Qi}
\cortext[cor1]{Corresponding author}
\cormark[1]
\ead{qzlhit@gmail.com}
\address[3]{School of Information Engineering, Guangdong Eco-Engineering Polytechnic, Guangzhou 510520, China}

\author[4]{Philip S. Yu}
\ead{psyu@uic.edu}
\address[4]{Department of Computer Science, University of Illinois Chicago, Chicago 60607, USA}

\title [mode = title]{Mixture of Experts (MoE): A Big Data Perspective}

\begin{abstract}
 As the era of big data arrives, traditional artificial intelligence algorithms have difficulty processing the demands of massive and diverse data. Mixture of experts (MoE) has shown excellent performance and broad application prospects. This paper provides an in-depth review and analysis of the latest progress in this field from multiple perspectives, including the basic principles, algorithmic models, key technical challenges, and application practices of MoE. First, we introduce the basic concept of MoE and its core idea and elaborate on its advantages over traditional single models. Then, we discuss the basic architecture of MoE and its main components, including the gating network, expert networks, and learning algorithms. Next, we review the applications of MoE in addressing key technical issues in big data. For each challenge, we provide specific MoE solutions and their innovations. Furthermore, we summarize the typical use cases of MoE in various application domains. This fully demonstrates the powerful capability of MoE in big data processing. We also analyze the advantages of MoE in big data environments. Finally, we explore the future development trends of MoE. We believe that MoE will become an important paradigm of artificial intelligence in the era of big data. In summary, this paper systematically elaborates on the principles, techniques, and applications of MoE in big data processing, providing theoretical and practical references to further promote the application of MoE in real scenarios.
\end{abstract}


\maketitle

\section{Introduction} \label{sec:introduction}

With the rapid development of information technology, the era of big data is arriving \cite{gan2023web,wan2024web3}. Massive, complex, and multisource big data have brought new opportunities and challenges to artificial intelligence (AI) \cite{winston1984artificial}. Compared with traditional single AI-based models, these big data have the following significant characteristics: 1) High-dimensional sparsity \cite{assent2012clustering}: Big data usually contain a large number of feature dimensions, but each sample only involves a small part of them, exhibiting high-dimensional sparsity. This poses a great challenge for modeling. 2) Heterogeneous multisource \cite{zhang2018multi}: Big data originate from various sensors, systems, and applications, presenting high heterogeneity and diversity. Effectively fusing these heterogeneous multisource data is a key problem that needs to be addressed. 3) Dynamic changes: Big data undergo continuous dynamic changes over time, requiring the ability to adapt quickly to environmental changes and achieve online learning and continuous optimization \cite{zhang2021adaptive}. 4) Complex relationships \cite{sun2022big}: Big data contain intricate and complex latent relationships that are difficult to fully and accurately capture with a single model. Therefore, the traditional single AI model has struggled to cope with the massive and diversified big data processing needs, making it urgent to explore new big data processing paradigms to overcome these challenges.

As a typical framework, the mixture of experts (MoE) \cite{jacobs1991adaptive,jordan1994hierarchical} has received considerable attention in recent years. It adopts a "divide-and-conquer" approach, dividing the complex learning task into multiple sub-tasks, which are handled by different expert networks, and a gating network dynamically coordinates and integrates the outputs of the experts to achieve effective modeling of complex big data. This divide-and-conquer approach has shown excellent performance in handling high-dimensional sparse data \cite{huang2024harder,liu2023uncovering}, fusing heterogeneous multisource data \cite{chenoctavius,wu2024lazarus}, achieving online learning \cite{kang2023real,mirus2019mixture}, and providing interpretability \cite{akrour2021continuous,germino2024fairmoe}. In recent years, MoE systems have attracted considerable attention in artificial intelligence. This neural network architecture \cite{bishop1994neural} uniquely integrates multiple models to achieve powerful problem-solving capabilities. Although the name contains the word "expert", no real experts are involved in the process. Therefore, how does the MoE system work? First, let's understand the basic principles of MoE. Traditional neural networks usually use only one model to handle all tasks, while MoE adopts a different strategy. Based on the different generation patterns of the data, MoE divides the dataset into multiple parts and trains an independent model for each part. These models are called "experts," and each of them is adept at handling a specific type of data. The gating module is responsible for selecting which expert to use based on the features of the input data. Finally, the model's output is a weighted combination of the outputs of the individual experts, with the weights determined by the gating module.

The distinctive feature of this paper is the comprehensive applications and technical challenges of MoE in big data processing. Although previous studies have reviewed the basic principles of MoE or its applications in specific fields, there is a lack of literature that delves deeply into the full-scale applications of MoE in big data processing. This paper aims to fill this gap. Through specific case studies, this paper demonstrates the practical application effects of MoE in fields such as natural language processing (NLP) \cite{hirschberg2015advances}, computer vision \cite{voulodimos2018deep}, recommendation systems \cite{jiang2023adamct}, and interdisciplinary areas under the big data context. This paper also analyzes how MoE overcomes traditional challenges in these domains. In addition, the paper not only emphasizes the advantages of MoE in the big data environment but also explores the challenges it faces, such as load imbalance and expert utilization. Based on this, the paper outlines the future development trends of MoE, further demonstrating the new application prospects of MoE in the big data era.

This comprehensive review aims to provide a thorough understanding of the advantages and innovations of MoE in big data processing and to offer theoretical and practical references for further promoting its application in real-world scenarios. The main contributions are as follows:

\begin{itemize}
    \item We systematically sort out the latest progress of MoE in the field of big data processing, starting from the basic principles, architecture, and key technical challenges of MoE (Section \ref{sec:introduction}). We provide a big data perspective for academic research and the industrial practice of MoE (Section \ref{sec:concepts}).

    \item We discuss in depth the key technologies of MoE and its solutions in the areas of high-dimensional sparse data modeling, heterogeneous multisource data fusion, real-time online learning, and interoperability (Section \ref{sec:technologies}). We also demonstrate its powerful capabilities and unique advantages in handling big data through typical cases in various application areas (Section \ref{sec:application}).

    \item We also analyze MoE's advantages of high scalability, efficient resource utilization, and better generalization capabilities in big data environments, and the challenges faced by MoE, such as load imbalance, expert utilization, gated network stability, and training difficulty (Section \ref{sec:advantages}). 

    \item  We further look forward to the future development trends of MoE, including the improvement of model generalization capabilities, the enhancement of algorithmic interpretability, and the advancement of system automation (Section \ref{sec:future}).
\end{itemize}

\section{Mixture of Experts} \label{sec:concepts}
\subsection{Basic Concepts and Development of MoE}

\begin{figure*}[ht]
    \centering
    \includegraphics[width=\linewidth]{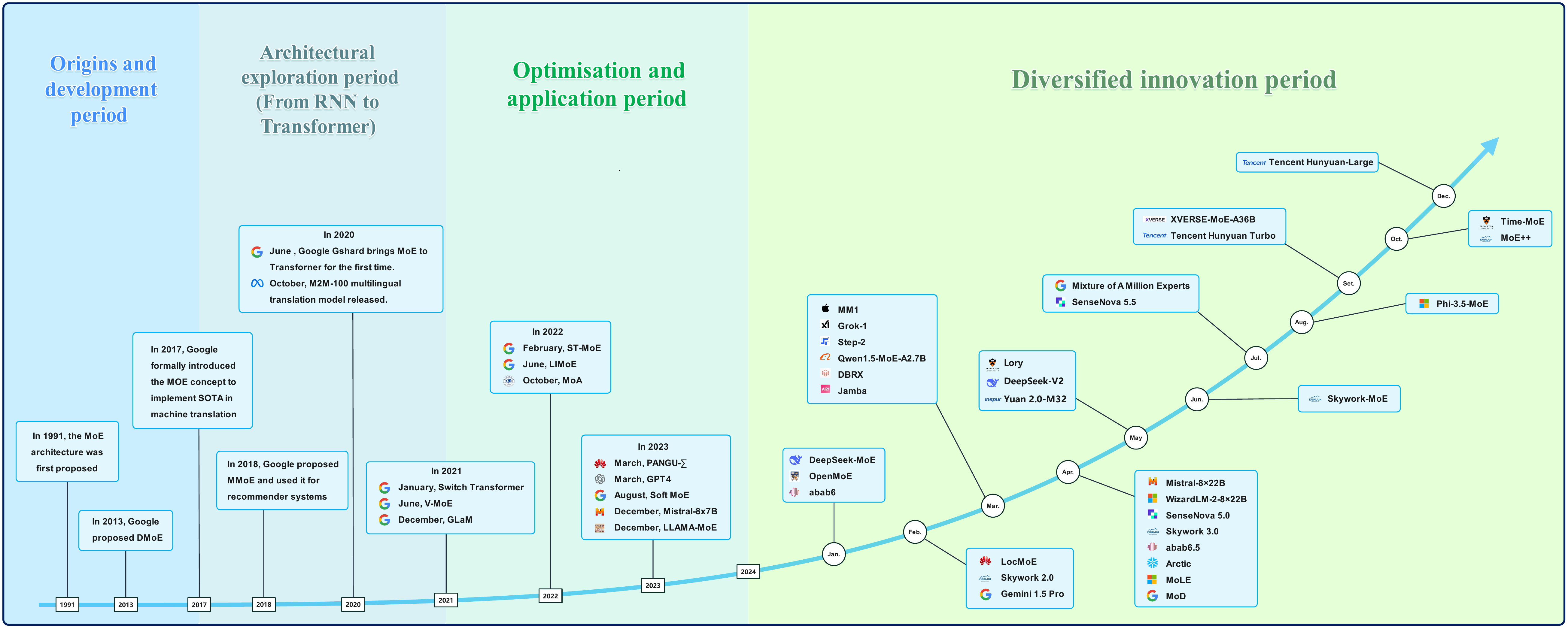}
    \caption{An overview of the chronological development of the MoE models.}
    \label{fig:timeline}
\end{figure*}

\textbf{Basic concepts}. The mixture of experts (MoE) model \cite{jacobs1991adaptive,jordan1994hierarchical} is based on the ensemble learning method and other processing algorithms. MoE brings together multiple specialized sub-models (i.e., "expert networks") to collaboratively handle complex tasks. Unlike traditional neural networks, the core idea of MoE is to divide the complex learning problem into multiple relatively smaller and simpler sub-tasks, and distribute the data to the most suitable experts (or learners). Each expert provides predictions based on the different characteristics of the input data in their respective areas of expertise. The Gating Network is responsible for dynamically selecting and integrating the outputs of these experts, ensuring that the system can make the optimal decision in different situations and efficiently solve complex tasks and challenges in big datasets. As shown in Figure \ref{fig:timeline}, it presents an overview of the chronological development of the MoE models.

\textbf{Origin and development of MoE}. In 1991, research led by Jacobs first proposed the mixture of experts (MoE) model \cite{jacobs1991adaptive}. The early model design was relatively simple, generally consisting of a small number of experts, each focusing on a specific data subset. Representative models in this stage include Jacobs' basic gating network and the hierarchical MoE (HME) \cite{jordan1994hierarchical}. These models achieved certain success in areas like speech recognition \cite{gales2006product}, demonstrating the effectiveness of MoE in handling specific tasks. However, as the scale of datasets expanded, the limitations of the early models gradually emerged, particularly in terms of computational resource utilization and model scalability. Over 20 years later, with the improvement of computational capability and the advancement of machine learning (ML) research \cite{mahesh2020machine}, the MoE architecture has undergone continuous improvements. In 2017, Google proposed the sparse MoE model \cite{shazeer2017outrageously}, which integrated MoE into long short-term memory (LSTM)  networks \cite{graves2012long}, creating a model with 137B parameters, using 10 times less computation than the previous dense LSTM models, and significantly improving the model's performance. The development of sparse gating mechanisms \cite{sabour2017dynamic} has also made MoE models more effective in selecting experts for complex tasks. However, although the MoE architecture has introduced new mechanisms and preliminarily demonstrated its potential for handling big data, it still faces challenges in terms of training time and computational resources in modern applications.

\textbf{Innovations in modern MoE architectures}. With the advent of the big data era, modern MoE architectures, especially their applications in big data processing and deep learning \cite{lecun2015deep} models, have gained new vitality. The new MoE architectures not only increase the number of experts but also introduce more efficient training methods and more complex gating mechanisms. In June 2020, Google released "GShard" \cite{lepikhin2020gshard}, which applied MoE to the Transformer model, replacing the feed-forward network (FFN) layers in the T5 (encoder-decoder) structure with MoE layers, and trained a series of models ranging from 12.5B to 600B parameters. Switch Transformers \cite{fedus2022switch} further simplified the routing strategy based on T5, training the largest model with up to 1.6 trillion parameters, and pushing the parameter scale of large language models (LLMs) \cite{gan2023model,zeng2023distributed} from hundreds of billions to trillions. In 2022, the ST-MoE model \cite{zoph2022st} was released, which conducted a deeper analysis of the structure and training strategies. In 2024, enterprises continuously released MoE-based LLMs, and the number of models with parameters exceeding 100 billion released from January to May alone surpassed the total of the previous three years. In addition to DeepSeekMoE \cite{dai2024deepseekmoe}, other companies released new MoE models (e.g., Databricks' DBRX \cite{gupta2024dbrx}, Alibaba's Qwen1.5-MoE-A2.7B \cite{yang2024qwen2}, Mistral AI's Mixtral 8$\times$7B \cite{jiang2024mixtral}), marking the continuous development of MoE in the field of LLMs. The recently released MoE++ architecture \cite{jin2024moe++}  has further innovated in performance by introducing the concept of "zero-compute experts", allowing each token to use a variable number of FFN experts, and enabling the simultaneous deployment of all experts on each GPU, unlocking greater performance potential than traditional MoE. Experimental results show that on LLMs with 0.6B to 7B parameters, MoE++ outperforms traditional MoE in terms of performance under the same model size while achieving 1.1 to 2.1 times the expert throughput speed. Through these developments, MoE has gradually become an important tool for processing big data and complex tasks.

\subsection{Basic Architecture and Mathematical Model}
\subsubsection{Architecture of MoE}

The MoE architecture has two main types: competitive MoE and cooperative MoE. In competitive MoE, data is forcibly partitioned into different discrete spaces, and each expert is responsible for handling a specific space. This architecture helps achieve specialization among models, but may also lead to overly rigid data space partitioning, which is not conducive to solving complex problems. In cooperative MoE, data partitioning is not strictly constrained, and the experts can jointly process the input data, enabling a more flexible problem-solving approach. The MoE structure is mainly composed of three core components: the gating mechanism, the expert network, and the output layer. Based on existing studies \cite{jordan1994hierarchical,shazeer2017outrageously}, details are described below.

\begin{figure}[ht]
    \centering
    \includegraphics[scale=0.23]{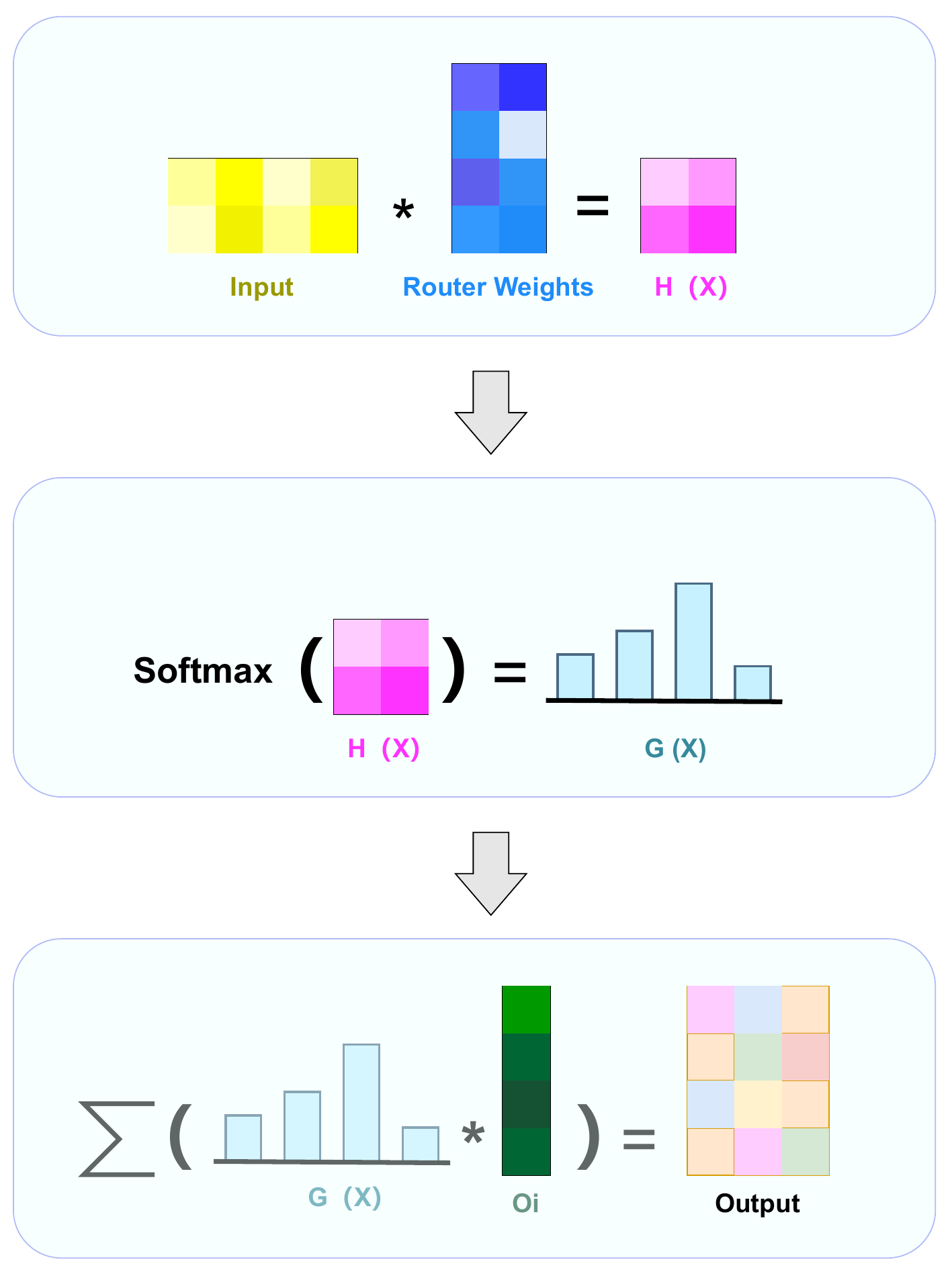}
    \caption{Gating mechanism for MoE model schematic.}
    \label{fig:gating}
\end{figure}

\textbf{Gating mechanism}: This mechanism is responsible for dynamically selecting the most suitable expert based on the input data. Modern gating mechanisms usually adopt a sparse gating strategy, where only a selected subset of experts is activated for each input sample during the forward pass. The Softmax gating function models the probability distribution over experts or tokens, and achieves sparsity by only computing the weighted sum of the outputs of the top $k$ experts. The mathematical model of the sparse gating strategy is as follows:

{\scriptsize
\begin{align*}
&1.  G(x)=\operatorname{Softmax}(\operatorname{KeepTopK}(H(x), k)) \\
&2.  H(x)_{i}=\left(x \cdot W_{g}\right)_{i}+\operatorname{StandardNormal}() \cdot \operatorname{Softplus}\left(\left(x \cdot W_{\text {noise }}\right)_{i}\right)\\
&3. \operatorname{KeepTopK}(v, k)_{i}=\left\{\begin{array}{ll}v_{i} & \text { , if } v_{i} \text { is in the top } k \text { elements of } v \\-\infty & \text { , otherwise }\end{array}\right.
\end{align*}
}
where: (1) $G(x)$: The output probability distribution of the gating network on input $x$. $H(x)$: The raw score function of $x$ after going through the gating network. \textit{KeepTopK(H(x),k)}: Retains the $k$ elements with the highest scores in $H(x)$. \textit{Softmax}: A function that converts a vector into a probability distribution. (2) $H(x)_i$: The score of $x$ on the $i$-th expert. \textit{StandardNormal}(): Random noise. \textit{Softplus}: A smooth non-linear activation function. (3) \textit{KeepTopK(v,k)$_i$}: The value of the $i$-th element in vector $v$ after the \textit{TopK} operation. $v_i$: The $i$-th element in vector $v$. $k$: The top $k$ expert subset.  Details are shown in Figure \ref{fig:gating}.

\begin{figure}[ht]
    \centering
    \includegraphics[scale=0.3]{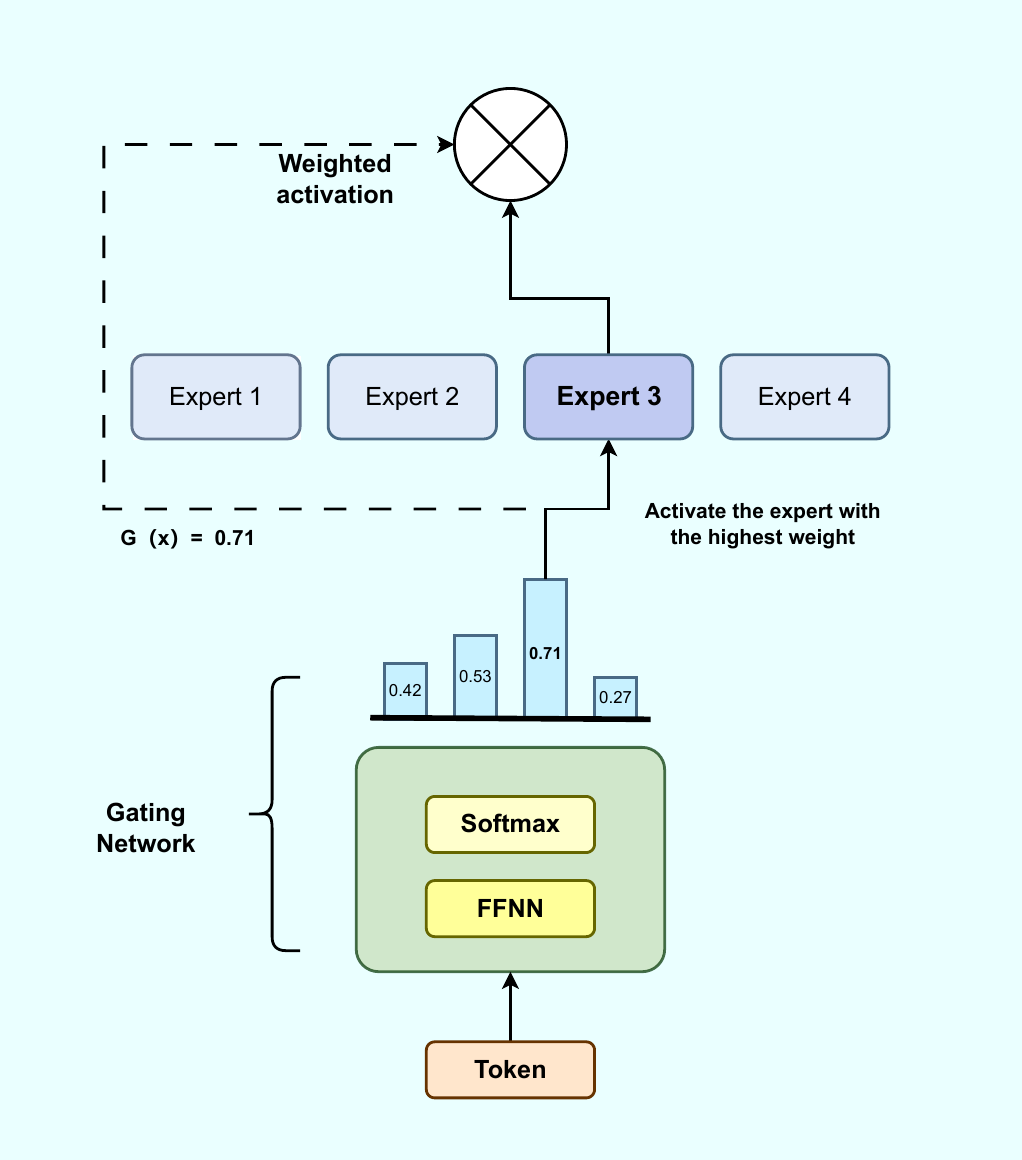}
    \caption{Expert network for the MoE model schematic.}
    \label{fig:expert}
\end{figure}

\textbf{Expert network}: The fundamental part of the MoE structure, composed of n independent neural networks. Each expert network can be an independent ML model, such as a neural network or a decision tree, focusing on its own features or tasks, with its parameters ${\theta }_{i}$. Each expert network ${f}_{i}(x;{\theta }_{i})$ can receive specific gating inputs and provide its decision output ${O}_{i}$. The outputs of the experts can be combined through weighted averaging to form the final result. Details are shown in Figure \ref{fig:expert}.

\textbf{Output layer}: Combines the outputs of the individual experts to produce the final prediction result. The computation of the output layer can be represented as:
$$
\begin{aligned}
\textit{FinalOutput} &= \sum  {_{i\in \textit{TopKIndices}}}{G(x)}_{i}\cdot {O}_{i}
\end{aligned}
$$

In specific applications, the MoE architecture can be flexibly designed according to the task requirements. For example, different expert network structures can be used, or a hierarchical gating network can be introduced. The key is to fully leverage the "divide-and-conquer" advantage of MoE to improve the modeling capability for complex big data.

\subsubsection{An Example of the Computational Method}

\begin{figure}[ht]
    \centering
    \includegraphics[scale=0.32]{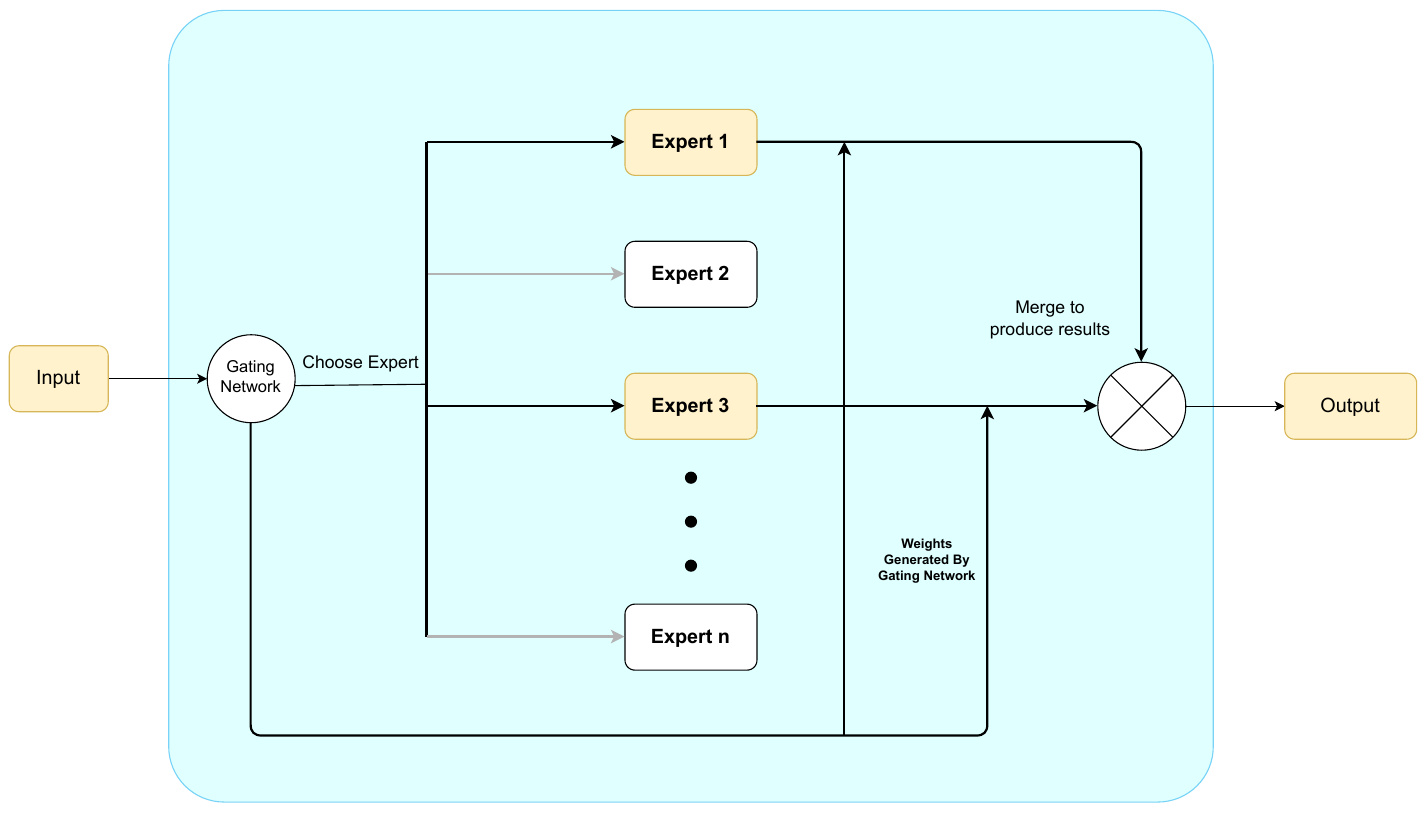}
    \caption{The simple schematic of how the MoE model works.}
    \label{fig:robot}
\end{figure}

The main difference between MoE and the traditional Transformer is that the MoE model can better handle complex and diverse tasks. Suppose we have a robot that can help us answer questions. A traditional Transformer is like an all-purpose robot that can answer many questions, but may not be as good at some special questions. The MoE model is like a group of robots, each of which is good at answering a certain type of question. When the MoE model encounters a question, it will let the robot that is good at that question answer it. As shown in Figure \ref{fig:robot}, the working principle of the MoE model is as follows: (1) First, the MoE model has many experts (robots), and each expert is good at handling a certain part of the problem. (2) Then, the MoE model has a network called the gating network, which can determine the weight of each expert. The weight represents the degree of trust the MoE model has in each expert. (3) When the MoE model encounters a new question, it will input the question into the gating network, which will tell the MoE model which experts it should trust. (4) The MoE model will let the selected experts handle the problem and combine their answers to get the final solution.

For easier understanding, here is a simplified example of an MoE model, Table \ref{calculation} shows the corresponding calculated data. Assumptions: A MoE model has three experts, each of which is an independent neural network that can process the input data $x$ and output a corresponding score. There is also a gating network that assigns weights to each expert, based on the ability of each expert to handle the input data. Steps: (1) Process the input: Each expert receives the same input data $x$ and outputs a score. (2) Gating mechanism assigns weights: The gating function evaluates the output of each expert and assigns a weight to each expert. (3) Select Top $K$ experts: Based on the weights assigned by the gating function, select the $K$ experts with the highest weights. In this example, we select the top two experts. (4) Weighted combination to form the final prediction: Multiply the selected expert outputs by their respective weights, and then sum these weighted outputs to form the final prediction result. 

\renewcommand{\arraystretch}{1.3}
\begin{table*}
\centering
\caption{A simplified example of MoE calculation}
\label{calculation}
\begin{tabular}{|c|c|c|c|c|c|} 
\hline
\textbf{Step} & \textbf{Expert number} & \begin{tabular}[c]{@{}l@{}}\textbf{Expert network} \\${f}_{i}(x;{\theta }_{i})$\end{tabular} & \begin{tabular}[c]{@{}l@{}}\textbf{Gating network} \\ $G(x)$ \end{tabular} & \begin{tabular}[c]{@{}l@{}}\textbf{Output of experts} \\${O}_{i}={f}_{i}(x;{\theta }_{i})$\end{tabular} & \begin{tabular}[c]{@{}l@{}}\textbf{Weighted output} \\${G(x)}_{i}*{O}_{i}$\end{tabular}  \\ 
\hline
1     & 1               & ${f}_{1}(x;{\theta }_{1})$                                                         & 0.5                                                           & 1.2                                                                                           & 0.6                                                                             \\ 
\hline
2     & 2               & ${f}_{2}(x;{\theta }_{2})$                                                         & 0.3                                                           & 0.8                                                                                           & 0.24                                                                            \\ 
\hline
3     & 3               & ${f}_{3}(x;{\theta }_{3})$                                                         & 0.2                                                           & 1                                                                                             & exclusion from calculations                                                     \\
\hline
\end{tabular}
\end{table*}

In Table \ref{calculation}, the final result is the sum of the weighted outputs, i.e., \textit{FinalOutput} = 0.5 $\times$ 1.2 + 0.3 $\times$ 0.8 = 0.84. In this way, the MoE model can flexibly combine the knowledge of multiple experts to improve the accuracy and robustness of the prediction. By utilizing the strengths of each expert, the MoE model can improve overall performance. MoE models are particularly effective when the problem space is large and complex, as they can capture various patterns and relationships. In contrast, the traditional Transformer model applies the same computation to all inputs without explicit specialization. While Transformer-based models have achieved remarkable success in many NLP tasks, they may be limited in handling more complex and diverse problems.

\subsection{Advantages}

Compared to traditional single ML models, MoE has the following significant advantages:
\begin{itemize}
    \item Enhanced modeling capability: By dividing a complex task into multiple sub-tasks, MoE can better capture local features and intricate relationships in the data, thereby improving overall modeling capability.

    \item Improved generalization performance: Different expert networks can focus on different subspaces of the processed data separately, thereby enhancing the model's generalization ability to new data.

    \item Increased computational efficiency: Since the expert networks can be trained and inferred in parallel, the computational efficiency of MoE is significantly higher than that of a single LLM.

    \item Enhanced interpretability: The structure of MoE makes the model's decision-making process more transparent, which is beneficial for improving the model's interpretability.
\end{itemize}

When dealing with massive big data, the core idea of MoE is to utilize the "divide-and-conquer" strategy to address various challenges in big data processing effectively:

\begin{itemize}
    \item High-dimensional sparse data modeling: By partitioning the original high-dimensional data into multiple relatively low-dimensional subspaces and having different expert networks perform specialized modeling, MoE can better capture the local structural information in the data, thereby improving the modeling accuracy.

    \item Heterogeneous multisource data fusion: Different expert networks can focus on processing heterogeneous data from different sources, and the gating network can dynamically combine the outputs of the experts to achieve effective fusion of complex data.

    \item Continuous learning, online learning, and continuous optimization: MoE can quickly adapt to environmental changes and achieve continuous online learning of dynamic big data by adding, removing, or fine-tuning expert networks.

    \item Improved model interpretability: The hierarchical structure of MoE makes the model's decision-making process more transparent, which is beneficial for explaining the model's behavioral logic and enhancing user trust in the model.
\end{itemize}

\section{Key Technologies of MoE for Big Data}  \label{sec:technologies}

Based on the core ideas mentioned above, MoE has demonstrated powerful capabilities in addressing key technical challenges in big data, effectively handling the diversity and complexity of data, and improving model performance and generalization ability. We now systematically explore these aspects in detail.

\subsection{High-dimensional Sparse Data Modeling}

High-dimensional sparse data \cite{assent2012clustering} is a typical characteristic of big data, where the feature dimensionality is very high in the dataset but the number of non-zero features in each sample is relatively small. This can lead to computational complexity, difficulties in handling sparsity, overfitting risks, and feature selection complexity, posing significant challenges for machine learning. MoE addresses the challenges of high-dimensional sparse data by decomposing it into multiple low-dimensional subspaces and having the expert networks learn them independently \cite{liu2023uncovering}. This approach effectively reduces the computational complexity, enhances the model's generalization ability \cite{huynh2019estimation}, and improves the capture of local data structure through dynamic routing \cite{huang2024harder}.

\textbf{Subspace partitioning based on local structure}. MoE first needs to perform effective subspace partitioning of the original high-dimensional data to capture the local data structure. Cluster analysis is the currently popular method \cite{assent2012clustering}. Cluster analysis is used to group the data points in the high-dimensional dataset into similar feature subsets or clusters, aggregating the data points with similar features into different subspaces. MoE can divide the high-dimensional data space into subspaces with different centers, considering the intra-group correlations within the different subspaces to achieve dimension reduction. By partitioning the original high-dimensional data into multiple relatively independent subspaces based on the local structural characteristics of the data, MoE can decouple the large-scale complex data into several simple low-dimensional tasks, better capturing the local structural information of the data and laying a solid foundation for subsequent modeling.

\textbf{Specialized modeling by expert networks}. For each subspace, MoE will train a specialized expert network for modeling. These expert networks can be various types of machine learning models, such as neural networks \cite{mcculloch1943logical} and decision trees \cite{quinlan1986induction}, and can choose the appropriate model structure based on the characteristics of the subspace to achieve optimal local feature learning. This specialized modeling approach not only better captures the local data structure but also avoids overfitting by reducing the complexity of the model on the entire dataset, further enhancing the robustness of the model on high-dimensional sparse data. With the help of these techniques, the model can more effectively learn useful features when processing high-dimensional sparse data while maintaining a relatively low risk of overfitting.

\textbf{Dynamic coordination by a gating network}. In traditional ML models, data processing usually relies on a global model, where all input samples use a single model architecture to fit the entire dataset. This static modeling approach has obvious limitations, especially in dealing with high-dimensional sparse data, as it cannot fully capture the complex heterogeneity and local features within the data, resulting in poor fitting of certain features and a lack of generalization ability. However, this is the advantage of the MoE gating network. When new sample data are input, the gating network dynamically calculates the weights of the expert networks and routes the input to the most suitable expert for prediction \cite{huang2024harder}. The flexibility of this dynamic coordination mechanism is significant in addressing high-dimensional sparse data, allowing MoE to fully utilize the specialized advantages of different expert networks and improve the modeling accuracy of high-dimensional sparse data. Through the "divide-and-conquer" strategy mentioned above, MoE has shown excellent performance in high-dimensional data modeling \cite{chamroukhi2019regularized}. Compared to a single LLM, MoE better captures the local features of the data, thereby significantly improving the fitting capability for complex high-dimensional data. For example, in spectral data calibration, previous studies used a sparse Bayesian MoE model to select the sparse features of the multivariate calibration model, effectively handling the high-dimensional sparse data and improving the accuracy and efficiency of spectral data calibration.

\subsection{Heterogeneous Multisource Data Fusion}

Heterogeneous data typically originates from various sensors, systems, and applications, exhibiting a high degree of heterogeneity and diversity. The differences in data format and structure increase the difficulty of model fusion and integration. How to effectively fuse these heterogeneous multisource data is another key technical challenge facing MoE in big data applications \cite{zhang2018multi}.

\textbf{Heterogeneous data preprocessing and encoding}. MoE first needs to perform unified preprocessing and encoding of heterogeneous data from different channels, unifying data from different sources and types to the same standard \cite{zhang2003data}. This includes: 1) Missing value processing: Big data may have missing values due to equipment failures, network interruptions, and other reasons during the data collection process. Using methods such as interpolation and deletion to handle missing values, ensuring data integrity, reducing prediction errors caused by missing values, and ensuring the temporal continuity and robustness of the data. 2) Feature normalization: The range of different feature values in big data can vary greatly, which may cause some features to dominate the model training, affecting the fairness and accuracy of the model. Therefore, it is necessary to normalize the data with different feature value ranges, eliminate the influence of feature value ranges, reduce the mutual influence between features, and make them comparable. 3) Categorical variable encoding: If no encoding is performed, the model will find it difficult to handle non-numerical data. Textual, categorical, and other data types in big data need to be converted to numerical data to facilitate ML models, improve the model's interpretability and generalization ability, reduce dimensionality, and enhance computational efficiency.

\textbf{Heterogeneous modeling of expert networks}. For different types of data sources, MoE will train dedicated expert networks for modeling \cite{zhou2024navigating,zhu2024task}. These experts can use different model structures to adapt to their data characteristics. For example, for image data, using convolutional neural networks (CNNs) \cite{li2021survey} can effectively extract spatial features and reduce the impact of mutual interference between tasks on network performance. For text data, Transformer-based models (such as BERT) \cite{devlin2018bert} can capture long-range dependencies and better understand complex text content \cite{hallee2024contrastive}. Furthermore, for time series data such as medical logs or stock prices, recurrent neural networks (RNNs) \cite{lipton2015critical} or long short-term memory (LSTMs) can better handle time dependence and have become popular and widely used models for predicting future events or results \cite{lee2022learning}. In this way, each expert network can delve into and understand the inherent characteristics of its data type.

\textbf{Adaptive fusion of gating networks}. When inputting new sample data, the gating network will adaptively adjust the weights of each expert network according to the characteristics of the data, realizing the dynamic fusion of heterogeneous multisource data \cite{wu2024lazarus}. This adaptive fusion mechanism allows MoE to fully utilize the strengths of each expert, improving the modeling effect on complex data. The innovative gating function effectively fuses data of multiple modalities, solving the problem of large differences in format and content, and retaining as many features of the multisource data as possible, improving the convergence speed of the fusion \cite{han2024fusemoe}.

\subsection{Real-time Online Learning}
 
Big data often exhibits dynamic changes, where the continuous growth of data volume and the dynamic changes in the data environment require models to have the ability to adapt quickly, enabling online learning and continuous optimization \cite{zhang2021adaptive}. The online learning capability of MoE is particularly important in an ever-evolving data environment. Leveraging its flexible model structure and advanced learning algorithms, MoE can demonstrate excellent online learning capabilities in addressing the dynamics and continuous optimization of big data, making it an ideal choice for handling dynamically changing big datasets.

\textbf{Online addition and deletion of expert networks}. MoE can dynamically add, remove \cite{huang2024harder}, or fine-tune \cite{chen2023sparse,liu2024moe,rypesc2024divide,shen2023mixture} expert networks in response to the immediate characteristics of the data, to adapt to changes in data distribution. When a new data subspace is detected, MoE can immediately train a corresponding new expert network to adapt to the new data subspace \cite{lee2024continual,li2020privacy}. For example, in the training of LLMs, MoE can effectively expand the model capacity while minimizing computational cost through a sparse gating mechanism \cite{nie2022hetumoe,rajbhandari2022deepspeed}, avoiding catastrophic forgetting, and smoothly adapting to changes in data distribution while maintaining previous knowledge \cite{chen2023lifelong}. Furthermore, when certain data types or patterns no longer appear frequently or the structure changes slightly, MoE can remove or fine-tune outdated expert networks to address the issues of high memory consumption and expert redundancy \cite{zadouri2023pushing}, ensuring the model maintains high efficiency, accuracy, and computational resource savings.

\textbf{Online optimization of gating networks}. Continuous optimization of the gating network is another key aspect of online learning. The online learning mechanism of the gating network allows the model to constantly self-adjust based on continuously updated data, ensuring optimal performance at different periods. MoE typically uses the expectation-maximization (EM) algorithm \cite{jordan1994hierarchical} or gradient descent \cite{akbari2023alternating,andrychowicz2016learning} to perform real-time adjustments of the gating network parameters and optimize the weight allocation of each expert network. It can also track changes in data distribution in real-time, efficiently capturing local data characteristics, and flexibly allocating computational resources for different subspaces to improve robustness. This dynamic coordination mechanism enables MoE to track changes in data distribution, quickly adjust the activation state of expert networks, and maintain prediction accuracy.

\textbf{Incremental learning algorithms}. MoE can also adopt incremental/continual learning algorithms \cite{gepperth2016incremental} to further improve online learning efficiency \cite{celik2022specializing,yu2024boosting}. Compared to traditional batch learning, incremental learning has a clear efficiency advantage. In the case of huge data volumes, retraining the entire model not only takes time and effort but also consumes a large amount of computational resources. Incremental learning algorithms allow the model to update only the relevant parameters when new data arrives, without retraining the entire network. This approach significantly reduces computational costs and is suitable for handling the continuous flow of big data. Incremental learning algorithms achieve model updates through the gradual learning of new data. They have the advantages of high efficiency, real-time, and strong adaptability compared to batch learning algorithms. Additionally, incremental learning avoids the forgetting effect when dealing with data changes, ensuring that the model does not lose important historical information due to new data \cite{yu2024boosting}.

By leveraging strategies such as flexible expert network structure adjustment, online gating optimization, and incremental learning, the MoE model demonstrates significant advantages and the ability to adapt quickly to environmental changes in big data fusion applications. It is ideal for handling dynamic big datasets, providing powerful solutions for various practical problems, and making it unparalleled in big data fusion applications \cite{gupta2024offline,kang2023real,mirus2019mixture}.

\subsection{Enhancing Model Interpretability}

Model interpretability determines the user's trust in its decisions and has a profound impact on the model's practical application in various domains. As a typical black-box model \cite{bunge1963general}, a single ML model often struggles to explain its internal decision-making process, which can affect the user's trust in the model's output. In the era of big data, the volume and complexity of data have greatly increased the demand for model interpretability, and the hierarchical structure of MoE has shown unique advantages in enhancing model interpretability. Through its gating network and multiple expert networks, MoE provides a highly interpretable structure that can further enhance user trust and promote its widespread deployment in big data applications \cite{germino2024fairmoe}.

\textbf{(1) Hierarchical structure provides interpretability}. The hierarchical structure of MoE, including the gating network and multiple expert networks, makes the entire model decision-making process more transparent. Users can understand how the input data is first routed to the appropriate expert networks, and how each expert network processes specific data subspaces to make key decisions that influence the final output \cite{akrour2021continuous}. Compared to traditional models, the hierarchical decision-making process of MoE appears more transparent, as traditional deep learning models often cannot provide sufficient explanations to cope with these diverse data \cite{minoura2021scmm}. The high-dimensional features and diverse data types in big data precisely require such a highly transparent model structure. This clear explanation path can trace the entire process of the model extracting features from data, allocating weights to expert networks to process data subspaces, and integrating the output, thereby 
enhancing user understanding and trust in the model.

\textbf{(2) Expert network-specific explanations}. Since each expert network is focused on processing a specific data subspace, its internal decision logic is often simpler and easier to explain. The complexity of big data often involves data features from different domains, such as medical imaging data and genetic sequence data, which have drastically different data characteristics. MoE's expert networks can cultivate the specialized capabilities of different experts, using more targeted and refined experts to handle these different complex datasets \cite{ismail2022interpretable,wang2022st}. Users can have higher confidence in the expert networks corresponding to specific domains. They can also optimize and debug the models based on the specific data characteristics. In comparison, traditional single models may not be able to handle such complex data distributions and features effectively, and the expert networks of the MoE model simplify the decision-making path, enhancing the overall model interpretability.

\textbf{(3) Gating network decision explanations}. As the core decision-making component of MoE, the gating network is responsible for routing input data to the various expert networks. By analyzing the internal parameters of the gating network, users can understand which features play a key role in the final decision \cite{jiang2024m4oe}. This clear feature weight analysis not only helps users understand the output contribution of each part of the model but also enables them to identify the most influential key factors in the data, thereby enhancing the model's transparency and trustworthiness. Another major advantage of the gating network is that its dynamic gating adjustments can also be understood in real-time by users. This dynamic selection mechanism not only endows the model with flexibility but also allows users to analyze the changing trend of weight at each time point, enabling users to understand the dynamic shifts in the importance of different data features and predict the model's behavior \cite{jiang2024m4oe}.

\textbf{(4) Human-computer interaction explanation mechanism}. To further enhance interpretability, MoE can also utilize a human-computer interaction (HCI) mechanism \cite{lin2023interaction}. In big data applications, user requirements and data characteristics are often dynamic and complex. Users can use a visualization and interactive interface to intuitively understand the internal logic of the model, and provide feedback to promote the continuous optimization and transparency of the model behavior \cite{luo2024towards,wang2024interpretable}. The human-computer interaction mechanism can help users better understand the model and propose modification suggestions during the model training and optimization process. Users can adjust the weights of specific features or introduce new expert networks to handle particular data subsets based on the model's interpretability results. Through this feedback loop mechanism, the MoE model can achieve self-improvement and gradually evolve towards a more transparent, interpretable, and user-expectation-compliant direction.

The hierarchical structure of MoE, the specialized decision-making paths of the expert networks, the feature weight explanation mechanism of the gating network, and the HCI system collectively construct a transparent and highly interpretable framework. The interpretability of the MoE model not only helps establish user trust, but also allows the model's decision-making process to be continuously optimized through user feedback, providing important support for the iterative upgrade of the model, and enhancing user trust in the model through the feedback loop. The hierarchical structure and expert network design of the MoE model make it highly interpretable, which is crucial for the deployment and optimization of the model in big data.

\subsection{Other Key Technologies}

In addition to the four major technical challenges mentioned above, MoE also involves the following key technologies of optimization strategies for big data applications:

\subsubsection{Optimized Design of Gating Networks}

In big data analysis, the traditional Softmax gating mechanism may not effectively identify the differences between different data types, which can impact the model's prediction accuracy. Previous studies have proposed various strategies to improve the interpretability of the gating algorithm.

\textbf{1) Attention mechanism} \cite{niu2021review}: The heterogeneity of big data means that data comes from multiple sources with different formats, structures, and statistical characteristics. In this context, a simple global weight allocation cannot satisfy the fine-grained processing requirements of different types of data. Therefore, using the attention mechanism to allocate weights to different types of data and optimize the scheduling strategy of different experts can enable more fine-grained decision-making, improving the learning efficiency of the gating network \cite{he2024mixture}. In recent years, gating network technologies based on the attention mechanism have made great progress. For example, the Yuan 2.0-M32 model introduces an attention-based gating network to construct an MoE with 32 experts \cite{wu2024yuan}. This mechanism focuses on the collaboration measure between expert models, effectively solving the problem of lacking association between two or more experts participating in the computation in traditional gating networks. This significantly improves the collaborative data processing among the experts, achieving higher model efficiency with less computational power. It is particularly suitable for processing resource-intensive tasks in big data scenarios, and enhances the scalability and resource utilization efficiency of the model in such applications.

\textbf{2) Reinforcement learning} \cite{kaelbling1996reinforcement}: It is a trial-and-error method aimed at software intelligent agents taking actions that maximize rewards in a specific environment. In a big data environment, the generation and changes of data are dynamic, so applications often need to adjust the model's decision-making strategy through learning. By introducing reinforcement learning, the gating network can self-learn and optimize decisions in a constantly changing environment. Through continuous trial-and-error and reward mechanisms, the decision-making capability of the gating network can be enhanced, improving the accuracy and efficiency of the question-answering system. The use of reinforcement learning to optimize the gating network parameters helps the gating network to constantly iterate and self-optimize in a complex and changing environment, enabling it to respond quickly to the complex big data environment \cite{zhou2023facilitating}.

\textbf{3) Adaptive gating}: Big data often has very high dimensions, and different data dimensions may have varying importance and relevance. The general fixed gating structure cannot flexibly cope with such complex data distribution changes. The adaptive gating mechanism analyzes the characteristics of the input data and the current state of the model, dynamically adjusting the structure and parameters of the gating network to adapt to the diverse changes in big data \cite{liu2024adamole}. This design is adjusted based on the feedback information during model training, realizing adaptive response to different task complexities, and providing a valuable direction for future research on adaptive expert selection mechanisms. This can potentially expand the range of optimizing model performance in various tasks, effectively enhancing the adaptability and robustness of the MoE model in handling dynamic big data scenarios.

\subsubsection{Heterogeneous Design of Expert Networks}

The datasets in big data applications are often composed of multiple heterogeneous subsets from different sources, such as images, text, sensor data, network logs, etc. To fully leverage the rich information in big data, the design of expert networks must be heterogeneous to flexibly handle the processing needs of different data subsets. The expert networks in MoE can use different types of ML models, such as neural networks and decision trees. A complex but critical task is how to choose the appropriate expert network structure based on the characteristics of the data. The following are some detailed tactics and methods:

\textbf{1) Model combination and ensemble learning} \cite{dong2020survey}: In big data applications, different data subsets are suitable for different ML models. Therefore, the expert networks in MoE can adopt different model combination strategies, and select the most appropriate model to handle specific sub-tasks based on the data characteristics. Through ensemble learning, the system can integrate the prediction results of multiple different models, thereby achieving good performance on multiple tasks. This heterogeneous model combination strategy greatly enhances the system's adaptability and results in better model fitting in heterogeneous big data scenarios \cite{li2024locmoe}.

\textbf{2) Cross-domain transfer learning} \cite{pan2009survey}: In multi-domain and multi-task big data scenarios, there are significant differences in data characteristics. However, through cross-domain transfer learning, the expert networks in MoE can transfer pre-trained models from other domains to the current task, leveraging the model knowledge from existing domains and reducing the dependence on new data \cite{yuan2025auformer}. Transfer learning can improve the adaptability and generalization of the model in new domains, especially when dealing with insufficient data in cross-domain big data scenarios, where data annotation is costly or difficult to obtain. It can also reduce the training time and computational resources required for the model in the target domain \cite{chang2022towards}.

\textbf{3) Adaptive expert selection}: Big data scenarios often involve large, sequential, and streaming datasets, such as social media and IoT data. To address this, MoE uses an adaptive expert selection mechanism to dynamically select the most appropriate expert network to process the current data characteristics \cite{huang2024dynamic}. When the data volume and data characteristics change rapidly over time, the adaptive expert selection mechanism can dynamically select the most suitable expert network based on the data characteristics and model status, maximizing the system's prediction accuracy and efficiency while ensuring real-time responsiveness \cite{jin2024moe++,lepikhin2020gshard,munezero2023dynamic}. This adaptive selection not only enhances the system's flexibility and model-fitting capability but also effectively reduces the waste of computational resources, making big data analysis more intelligent and efficient.

\subsubsection{Deployment and Engineering Implementation}

One major drawback of the MoE model in addressing big data problems is its large parameter size, which becomes particularly evident in small-scale applications that require local execution. Successfully deploying the MoE model in real-world application scenarios requires solving a series of engineering implementation issues, such as model compression and hardware acceleration \cite{deng2020model}, to ensure the system's real-time performance and scalability. These issues become even more critical in the context of big data.

\textbf{1) Model compression}: In big data applications, the computational and storage costs of MoE can be very high. Using model compression techniques, such as pruning \cite{chen2022task,kim2021scalable,liu2024efficient}, quantization \cite{team2024jamba}, and distillation \cite{xu2024using}, can reduce the model size and lower the deployment costs. Switch Transformers \cite{fedus2022switch} explores the use of distillation techniques to convert the MoE model into a more compact and dense model, achieving a sparsity gain of about 30\% to 40\%, which enables the effective deployment of smaller-scale models in production environments. Inspired by the latest research results on LLMs and MoE routing strategies, the MoE-Pruner \cite{xie2024moe} model has achieved breakthroughs in model compression for LLMs. Compared to Switch Transformers, MoE-Pruner provides higher sparsity gains and does not require retraining the model during the pruning process. Additionally, through expert-level knowledge distillation, MoE-Pruner can further optimize the performance of the pruned model.

\textbf{2) Hardware acceleration}: To meet real-time requirements, several studies have also offloaded some model inference tasks to specialized hardware accelerators, such as GPUs \cite{he2021fastmoe,jin2024moe++,kamahori2024fiddler,pan2024parm} or FPGAs \cite{fan2022m3vit,sarkar2023edge}. These hardware acceleration solutions can significantly improve the inference speed of the MoE model, allowing it to maintain high efficiency when processing big datasets.

\textbf{3) Subnetwork extraction and weight merging}: Furthermore, "subnetwork extraction" \cite{almeldein2023accelerating} has emerged as a new technical approach, which allows extracting more targeted small subnetworks from large MoE architectures to adapt to different task requirements. Meanwhile, "weight merging" is an effective way to reduce the overall parameter size by integrating the weights of multiple expert nodes. Both of these techniques can help reduce computational complexity and parameter size \cite{tang2024merging}.

In summary, when considering the various key technical challenges in the big data era, MoE has great potential to leverage its advantages but requires further innovation in many aspects.

\begin{figure}[ht]
    \centering
    \includegraphics[scale=0.08]{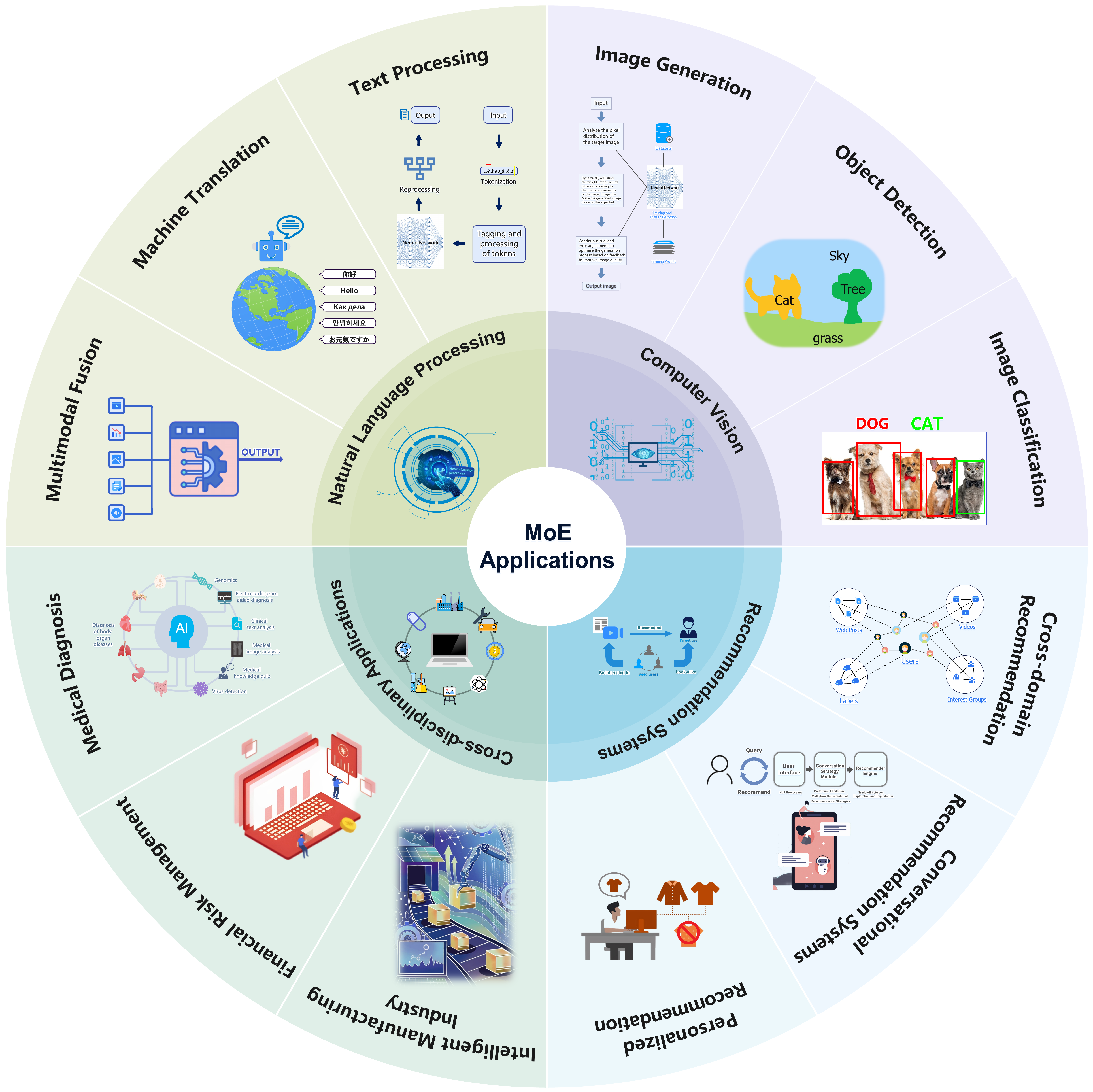}
    \caption{Typical use cases of the MoE model for big data.}
    \label{fig:application}
\end{figure}

\section{Typical Application Domain Cases} \label{sec:application}

Based on the key technologies mentioned above, MoE has achieved remarkable results in many big data application domains, effectively allocating resources and dynamically activating appropriate expert units. This has significantly improved the efficiency and accuracy of the model in processing high-dimensional and big data, demonstrating its excellent performance. Figure \ref{fig:application} illustrates several typical application cases of MoE in the context of big data. Below, we will introduce each of these applications in detail.

\subsection{Natural Language Processing}

Natural language processing (NLP) \cite{hirschberg2015advances} is one of the most widely applied domains of MoE. In LLMs, NLP tasks often involve high-dimensional data, complex semantic relationships, and dynamic language characteristics. Traditional models frequently face the dilemma of balancing efficiency and accuracy when dealing with big data. The MoE model has made significant progress in many classic tasks, especially in text processing, machine translation, and multimodal fusion, through its sparse activation mechanism. Since Google's pioneering work in 2017 on integrating MoE into LSTM layers \cite{shazeer2017outrageously}, the MoE model has quickly become an important technology in NLP research. As the data scale expands, the application of MoE in NLP continues to expand, and it has shown strong potential and adaptability.

\subsubsection{Text Processing}

Various MoE models are widely used to improve the efficiency and accuracy of text processing. The application of MoE models has consistently broken through the performance bottlenecks of traditional methods. To overcome the challenges of improving text generation quality and complex information processing in big data, many MoE-based methods were developed. Chai et al. \cite{chai2023improved} adopts a multi-generator approach and features a statistical alignment paradigm to enhance the representation capability of the generator in the MoE framework, improving the quality of adversarial text generation. This enhancement not only improves the fluency of the generated text but also improves the effectiveness of adversarial training in text-generation tasks. This model has brought breakthroughs in text generation, especially in handling highly complex text. For the logical table-to-text generation task, the LogicMoE model \cite{wu2024enhancing} enhances the semantic richness and logical fidelity of the generated sentences, successfully surpassing strong baseline models. It also enhances the diversity and controllability of the generated content. This capability is equally important for processing complex information and generating high-quality text. Another important advancement is the RetGen framework \cite{zhang2022retgen}, which jointly trains text generators and document retrievers, and integrates MoE, effectively improving the information content and relevance of the generated text, mitigating the data limitation and hallucinated facts of traditional models in dealing with big data. This approach, by combining text generation and document retrieval, makes the generated text more aligned with user needs and context information. Meanwhile, the MoE integration further enhances the flexibility and scalability of the model. In contrast, the QMoE model \cite{frantar2023qmoe} focuses on reducing the inference cost of LLMs in NLP, which is an important progress in MoE model quantization, reducing the inference cost of LLMs in NLP tasks, making high-performance text processing on cost-effective hardware possible, not only lowering the usage threshold of LLMs in NLP tasks but also solving the cost problem of processing big text data.

\subsubsection{Machine Translation}

In the field of machine translation, the application of the MoE model can greatly improve the quality of multilingual translation, particularly in the context of big data and multilingual situations, leading to better translation results. Multilingual translation systems need to handle the complex mapping relationships between different languages, especially in unbalanced resource situations, and traditional models often struggle to achieve a good balance between low-resource and high-resource languages. The MoE model, by integrating multiple expert networks, can effectively handle the conversion between different languages, improving the accuracy and fluency of the translation.

In 2022, the computational model based on sparsely gated MoE \cite{costa2022no} significantly narrowed the performance gap between low-resource and high-resource languages, covering more than 40,000 different language pairs in the Flores-200 benchmark dataset for human translation, achieving a 44\% BLEU improvement and driving the development of universal translation systems. A conditional computation model \cite{nllb2024scaling}, also based on a sparsely gated MoE, which extends the neural machine translation system to 200 languages using cross-lingual migration learning based on the study \cite{costa2022no}, significantly improves multilingual translation quality and language coverage. To address the problem of low inference efficiency of MoE models, Huang et al. \cite{huang2023towards} proposed three optimization techniques (dynamic gating, expert buffering, and load balancing) to enhance the inference efficiency of the MoE model in language modeling and machine translation tasks, reducing execution time and memory usage.

It should be noted that although the MoE model can theoretically be extended to handle a large number of languages and tasks, its actual deployment and operation require a large amount of computational resources and costs, limiting its widespread adoption in practical applications. Additionally, current research has mainly focused on high-resource languages, with inadequate support for low-resource languages, exacerbating the digital divide. Ensuring that open-source resources and technologies are widely accessible and usable, especially in low-resource language communities, remains a challenge \cite{costa2022no,nllb2024scaling}. Through continued research and optimization, these issues are expected to be gradually resolved, driving the development of machine translation technology towards a more universal and inclusive direction.

\subsubsection{Multimodal Fusion}

With the widespread use of new formats of big data information such as social media and video-sharing platforms, multimodal information has become an indispensable part of daily life. Multimodal NLP involves processing various types of big data, such as text, images, and audio, which places higher demands on existing ML models. Therefore, some studies focused on how to effectively combine information from different modalities to achieve a more comprehensive and accurate understanding.

In multimodal NLP tasks, the MoE model, with the diversity of its expert system, can effectively extract and fuse features from different modalities and handle information from them, enabling rapid development and widespread application in the face of ever-increasing data volume and complexity. In 2022, LIMoE \cite{mustafa2022multimodal}, as the first multimodal model to apply sparse MoE model, significantly improved the performance of multimodal learning through entropy regularization optimization of expert utilization, achieving competitive accuracy with advanced dense models. LIMoE has competed with advanced dense models in multiple multimodal tasks and achieved excellent performance in several benchmark tests, demonstrating the potential of the sparse activation mechanism in multimodal learning. Based on LIMoE, M3SRec 
\cite{bian2023multi} further applies the sparse MoE model to multimodal recommendation systems, significantly enhancing the modeling capability of complex user intent through deep cross-modal semantic fusion, especially in data-scarce situations. However, its limitation lies in the lack of stable generalization ability when the data volume is large or the modal differences are significant. In 2024, the Mistral 8$\times$7B model \cite{jiang2024mixtral} was released. This MoE model is composed of a combination of 70 billion-parameter small models, which outperformed the 70 billion-parameter LLaMA 2 in multiple performance metrics, particularly in code and mathematical benchmarks. Compared to the methods followed by other LLMs, the MoE-based architecture of Mistral 8x7B has advantages in multimodal processing. Table \ref{openMoE} shows the main performance benchmarks of some recent open-source hybrid expert models. Chinese MoE-oriented LLMs have made significant progress, with Tencent Hunyuan \cite{sun2024hunyuan} as China's first self-developed multimodal LLMs using the MoE architecture. It ranked first in the Chinese multimodal LLMs \cite{wu2023multimodal} with the SuperCLUE-V benchmark, demonstrating the strong capability of MoE in Chinese multimodal understanding.

\renewcommand{\arraystretch}{1.5}
\begin{table*}[!ht]
\centering
\footnotesize
\caption{Recent open-source hybrid expert models detailing some parameters, as well as performance benchmarks such as MMLU, C-Eval, CMMLU, GSM8K, MATH, and HumanEval, unless otherwise noted.}
\label{openMoE}
\begin{tabular}{l|c|c|cccccc} 
\hline
\multirow{2}{*}{\textbf{Name}} & \multicolumn{1}{l|}{\multirow{2}{*}{\textbf{Time}}} & \multicolumn{1}{l|}{\multirow{2}{*}{\textbf{Total params}}} & \multicolumn{6}{c}{\textbf{Benchmarks}}                    \\ 
\cline{4-9}
                      & \multicolumn{1}{l|}{}                      & \multicolumn{1}{l|}{}                              & \textbf{MMLU} & \textbf{C-Eval} & \textbf{CMMLU} & \textbf{GSM8K} & \textbf{MATH} & \textbf{HumanEval}  \\ 
\hline
Grok-1-A85B           & 2024.03                                    & 314B                                               & 73   & -      & -     & 62.9  & 23.9 & 63.2       \\
DBRX-A36B             & 2024.03                                    & 132B                                               & 73.7 & 44.9   & 61.3  & 70.7  & 25.6 & 46.3       \\
Qwen1.5-MoE-A2.7B     & 2024.03                                    & 14.3B                                              & 62.5 & -      & -     & 61.5  & -    & 34.2       \\
Mixtral-8x7B          & 2024.04                                    & 46.7B                                              & 70.6 & -      & -     & 77.4  & 28.4 & 40.2       \\
DeepSeek-V2-A21B      & 2024.05                                    & 236B                                               & 78.5 & 81.7   & 84    & 79.2  & 43.6 & 48.8       \\
Yuan 2.0-M32          & 2024.05                                    & 40B                                                & 72.2 & -      & -     & 92.7  & 55.9 & 74.4       \\
Skywork-MoE-A22B      & 2024.06                                    & 146B                                               & 77.4 & 82.2   & 79.5  & 76.1  & 31.9 & 43.9       \\
Jamba-1.5-Large       & 2024.08                                    & 398B                                               & 80   & -      & -     & 87    & -    & 71.3       \\
XVERSE-MoE-A36B       & 2024.09                                    & 255B                                               & 80.8 & 79.5   & 81.7  & 89.5  & 53.3 & 51.8       \\
Hunyuan-Large         & 2024.11                                    & 389B                                               & 88.4 & 91.9   & 90.2  & 92.8  & 69.8 & 71.4       \\
\hline
\end{tabular}
\end{table*}

When building a general multimodal LLM, fine-tuning various image-text instruction data is a crucial step. Due to the heterogeneity of big data, different instruction data configurations may lead to poor model performance on specific domain tasks. To address this issue, the LLaVA-MoLE model \cite{chen2024llava} extends the popular low-rank adaptation (LoRA) \cite{hu2021lora} method in the Transformer layers and creates a set of LoRA experts specifically for the MLP layers. It uses a routing function to direct each token to the most relevant top-k experts, allowing for adaptive selection of tokens from different domains, mitigating the cross-domain data conflicts in instruction fine-tuning of multimodal LLMs \cite{wu2023multimodal}, and achieving significant performance gains.

\subsubsection{Other Advancements}

In the direction of open-domain question answering, the generalist language model (GLaM) \cite{du2022glam} series has expanded the model capacity, with training costs only one-third of GPT-3, while simultaneously showing better zero-sample and one-time performance on 29 NLP tasks. This effective resource-saving allows the GLaM series to handle big data while accommodating even larger data volumes, showcasing the potential of this model in reducing resource consumption in language-based applications. For dialogue systems, the DeepSeek-V2 \cite{liu2024deepseek}, based on multi-head latent attention (MLA) and the DeepSeekMoE architecture, has reduced training and inference costs by 42.5\%. According to the evaluation results, even with only 21B activation parameters, DeepSeek-V2 and its chatbot version still achieve outstanding performance among open-source models. In visual language tasks, the Qwen-VL series \cite{bai2023qwen}, through innovative visual sensors and a multi-stage training pipeline, combined with a multilingual multimodal clean corpus containing rich image, video, and audio data, enables the model to better understand and process multimodal information. This ensures the consistency of the training data, avoids the impact of noisy data on model performance, and endows the model with powerful visual understanding capabilities. Qwen-VL and Qwen-VL-Chat have achieved excellent performance in tasks such as image description, question answering, and visual grounding, particularly outperforming existing visual language chatbots in real-world dialogue benchmarks.

\subsection{Computer Vision}

In the field of computer vision \cite{voulodimos2018deep}, traditional models mainly rely on deep learning techniques, such as convolutional neural networks (CNNs), which can effectively extract image features and perform classification. However, this traditional approach often struggles to fully leverage sparsity when faced with complex tasks. With the exponential growth in data scale, increasing dimensionality, and the dramatically growing volume of data that computer vision systems must process, higher demands are placed on the performance, efficiency, and generalization capabilities of the models. Some studies have recognized that the high dependence of traditional models on feature selection makes these methods often inadequate when facing complex visual tasks, and have consequently introduced the MoE model. The key to this model lies in its flexibility and dynamic computational capability. By leveraging its "divide-and-conquer" strategy, it effectively addresses the challenges of high dimensionality, locality, and complex semantics in visual data, becoming an important tool for handling various visual tasks in big data environments.

\subsubsection{Image Classification}

Image classification is one of the fundamental tasks in computer vision, intending to assign input images to predefined categories. In image classification tasks, MoE can assign images of different categories to specialized expert networks for modeling, and the gating network can dynamically combine the expert outputs to improve classification accuracy. Moreover, MoE can combine different models and features to enhance the accuracy and efficiency of classification, with significant advantages in handling complex image datasets. Machine learning methods have shown that this field also faces challenges related to data increments and insufficient model generalization capability in image processing and classification \cite{mo2022survey}, prompting the development of more efficient algorithms to improve image classification accuracy. The sparse V-MoE model \cite{riquelme2021scaling} can achieve higher classification accuracy in image recognition. This model further applies the MoE architecture to the Transformer-based computer vision models, demonstrating performance on par with top-performing dense networks in image classification tasks. The model scales the visual model up to 1.5 billion parameters, demonstrating the feasibility and efficiency of the sparse MoE model for numerous image recognition tasks, maintaining high accuracy while targeting big data. To address the low classification accuracy in small-sample remote sensing image scenarios, a transfer learning-based MoE (TLMoE) classification model \cite{gong2021transfer} fuses global and local features to achieve more accurate scene image classification. This not only improves the classification accuracy of small-sample datasets but also has the potential to be further extended to image classification tasks in big data environments. In contrast to the transfer learning-based TLMOE, a meta-learning-based MoE architecture \cite{vogiatzis2022novel} for multi-class image classification was proposed. It decomposes the task into smaller parts, utilizes the expert mixture scheme and Bayesian rules to reduce training complexity, further improving efficiency and accuracy, and eliminating the redundant learning process in large data, significantly reducing the number of operations in the training stage, with a significant improvement in accuracy over existing traditional image classification techniques. To address the problem of similar appearance between classes in fine-grained image classification, Zhang et al. \cite{zhang2021enhancing} proposed an enhanced expert mixture-based fine-grained image classification method, which leverages visual attention mechanisms and a gradually enhanced learning strategy to encourage experts to learn more diverse and discriminative representations, achieving excellent performance in weakly supervised localization and fine-grained image classification tasks. 

\subsubsection{Object Detection}

As a core task in computer vision, the accuracy and efficiency of object detection directly affect the performance of many real-world applications, such as self-driving, intelligent surveillance, and medical image analysis. In the context of big data, traditional object detection methods have gradually shown limitations in handling complex scenarios and diverse data. MoE can train expert networks for different types of targets, and the gating network can adaptively select the appropriate experts for detection based on the input image, successfully addressing the dual challenges of efficiency and accuracy in object detection with big data. In 2023, Zong et al. \cite{zong2023detrs} proposed a hybrid training strategy that successfully broke through the upper limit of many object detection models, setting new state-of-the-art results on COCO and LVIS, demonstrating the potential of MoE in handling many visual tasks. Compared to previous training strategies, SenseTime's AlignDet framework optimizes resource allocation through self-supervised pre-training, exhibiting strong universality across various detectors, making it an ideal choice for efficient detection applications in industrial big data scenarios. Also in 2023, Mod-Squad \cite{chen2023mod} integrates the MoE layer into the vision Transformer model and introduces a new loss function to encourage sparse but strong dependence between experts and tasks. This model is modularized into expert squads, matching experts and tasks during training to achieve cooperation and specialization in handling multiple visual tasks within a single model. This allows the extraction of lightweight sub-models targeted for object detection while maintaining the performance of the large model, enabling the selection of appropriate sub-models for specific object detection tasks based on the actual application scenario. In contrast, the AdaMV-MoE framework \cite{chen2023adamv} adaptively determines the number of experts activated for each task, avoiding the cumbersome process of manually adjusting model size. AdaMV-MoE is not only applicable to ImageNet classification but can also automatically determine the required number of experts for COCO object detection and instance segmentation tasks, enabling efficient multi-task visual recognition in big data environments. Unlike traditional models that require manual parameter and expert number adjustments, AdaMV-MoE reduces user intervention and simplifies the model optimization process. Its adaptive mechanism effectively improves the model's performance in various object detection tasks while reducing the dependence on large-scale computational resources. The significant progress of MoE technology in improving the accuracy, efficiency, and generalization of object detection foreshadows its widespread application prospects in the future field of computer vision.

\subsubsection{Image Generation}

In the process of image generation, models often face a dilemma between diversity and quality: traditional generative models tend to generate samples that are similar to the training data, resulting in a lack of diversity in the generated images. In addition, the generated images tend to have some unrealistic or blurred parts, which reduces the quality of the generated images. This dilemma becomes more prominent when dealing with complex scenarios or tasks with high-detail requirements. For such complex image generation tasks, MoE can assign different types of images to expert networks to handle the rich details, and the gating network can dynamically combine the expert outputs to ensure quality in big data scenarios, generating more realistic and diverse images. For example, the SDXL 1.0 \cite{sdxl} released by the Stable AI team uses a MoE latent diffusion process for text-to-image synthesis. SDXL not only outperforms the previous Stable Diffusion-based models \cite{podell2023sdxl} in generation quality but also achieves results comparable to the current black-box image generators, with good performance in image generation quality, control capabilities, multimodal capabilities, and open-source characteristics. In terms of specific implementation and application scenarios, the SDXL model focuses on generating high-resolution images, while another text-to-image generation model, mixture-of-attention (MoA) \cite{wang2024moa}, emphasizes the match between image quality and the fineness of the generated images. The MoA model is a novel personalized text-to-image diffusion model architecture that uses an innovative routing mechanism to allocate workloads between personalized and non-personalized branches. It has the advantage of controlling the subject and background in the generated images more precisely, optimizing the image generation process and allowing for more detailed subject-context control, thereby creating high-quality and diverse personalized images. As the data volume in various fields continues to grow, the application scope of MoE is also broadening, spanning many interdisciplinary areas, with an increasing demand for image generation across different disciplines. In artistic image generation, RAPHAEL \cite{xue2024raphael} combines a spatio-temporal MoE model to generate highly artistic images, accurately depicting multiple nouns, adjectives, and verbs in the text prompts of large datasets, and exhibiting excellent image quality and aesthetic appeal, marking the success of MoE models in handling complex image generation tasks in the artistic domain. Furthermore, in the field of fashion design, Text2Human \cite{jiang2022text2human} text-driven framework utilizes a hierarchical texture perception codebook and MoE technology to generate high-quality, diverse, and controllable human body images and achieves fine-grained control of clothing shape and texture through text input.

\subsubsection{Other Applications}

The highly scalable MoE stack Flex \cite{hwang2023tutel}, designed by Hwang et al., significantly improves the training and inference efficiency and accuracy of large MoE models in computer vision tasks through dynamic adaptive parallelism and pipelining techniques, accelerating the training and inference speed of the MoE-based SwinV2 \cite{liu2022swin} and enhancing the accuracy in pre-training and downstream computer vision tasks. In terms of adversarial robustness, adversarial attacks have always been a major challenge facing deep learning. Zhang et al. \cite{zhang2023robust} utilized the AdvMoE training framework, alternately optimizing the router and expert adversarial robustness, achieving a 1\% to 4\% improvement in adversarial robustness over the original dense CNN while reducing the inference cost by more than 50\%. The pMoE (patch-level MoE) model \cite{chowdhury2023patch}, which partitions the image into multiple patches and routes them based on priority, has made significant contributions in addressing the problems of computational resource tightness and lack of generalization capability in adversarial large-scale image datasets, reducing the computational cost and improving the model's generalization ability, driving technological progress in handling large-scale image datasets. For image fusion, the task-customized adaptive mixture framework \cite{zhu2024task} introduces shared and mutual information regularization constraints on expert adapters, enabling efficient cross-task image fusion in a unified model and ensuring that the characteristics of each task can be fully explored and utilized.

\subsection{Recommendation Systems}

At the forefront of the intersection between big data information retrieval and artificial intelligence, recommendation systems \cite{chen2024data} have become increasingly important as a bridge connecting users to big data information in domains such as e-commerce, social media, and content distribution platforms. They serve as a key factor in enhancing user experience and strengthening business competitiveness. However, recommendation algorithms still face many challenges when dealing with high-dimensional, dynamic, and diverse data, such as data sparsity, cold-start problems, and user interest drift. The unique structure of the MoE model is well-suited to effectively alleviate these issues. MoE can analyze large amounts of user behavior data to accurately identify the dynamic changes in user interests, optimize the interactions between experts, and further improve the model's adaptability to user interest changes, enabling recommendation systems to continuously learn and optimize in dynamic environments. This has garnered increasing attention in the fields of big data and artificial intelligence.

\subsubsection{Personalized Recommendation}

For personalized recommendation tasks, the choice of different recommendation methods is crucial to the effectiveness of the research results. The core objective of personalized recommendation is to provide content or product recommendations that meet the individual needs of users based on their historical behavior and current preferences. Traditional personalized recommendation methods, such as collaborative filtering and content-based recommendation, have shown certain limitations in handling issues like data sparsity \cite{chu2020leveraging}. Therefore, research on incorporating new models like MoE from deep learning into personalized recommendations has become increasingly prevalent. For personalized recommendation tasks, MoE can train expert networks for different types of user interests, leveraging multiple data sources to enhance the system's adaptability to user behavior changes and improve the performance and personalization of recommendation systems to address the challenges of big data.

In recent years, deep learning-based methods have been widely applied in personalized recommendations. Some studies have found that combining MoE with deep learning can capture users' short-term and long-term preferences as well as complex user features, improving the accuracy of recommendation systems in dynamic environments. For example, the study \cite{liu2023extending} made early attempts to incorporate MoE into deep learning-based personalized recommendation. This research extended the MoE framework to non-linear trajectory analysis, focusing on the heterogeneity within and between clusters. This approach considers the influence of covariates on individual differences, capturing the complexity and dynamism of user behavior. Perera combined deep learning to use covariates as clustering predictors, not only identifying latent classes but also explaining the patterns of user behavior change over time, providing a deeper understanding of personalized recommendation. Although this method has a higher computational complexity, it can provide more accurate personalized recommendations for user groups with highly complex behavioral characteristics. In contrast, a sequential recommendation (SR) model \cite{jiang2023adamct} based on the Transformer global attention mechanism and local convolutional filters, combined with the use of perception-adaptive MoE units and squeeze-excitation attention mechanisms, can capture users' long-term and short-term preferences while also being able to flexibly respond to users' dynamic needs. This allows for better understanding and prediction of user interest changes, providing more diverse and enriched recommendation content. Compared to traditional collaborative filtering and content-based recommendation methods, deep learning methods combined with MoE can not only handle high-dimensional sparse data but also solve the problem of traditional sequential recommendation algorithms, which often struggle to simultaneously consider users' long-term history and short-term needs through large-scale parallel computing.

The research by Lai et al.  \cite{laisong2023next} further expanded the multi-task personalized recommendation scenario of the MoE framework. Their sparse-sharing multi-gate MoE next-interest recommendation algorithm uses multi-task learning and sparse-sharing structures to generate corresponding expert networks for different tasks. By optimizing expert selection through a deep learning-based iterative magnitude pruning method, this approach effectively reduces model complexity while improving the accuracy of personalized recommendations and the performance of multi-objective tasks. Compared to the SR model, the sparse-sharing multi-gate MoE model focuses more on reducing computational overhead through structural optimization, while maintaining high recommendation accuracy and enhancing the model's scalability, making it suitable for scenarios with complex and heterogeneous user behavior. To address the issue of confounding features in personalized recommendation, a causal inference-based framework, called deconfounding causal recommendation (DCR) \cite{he2023addressing}, trains separate expert modules to manage confounding features in recommendations, reducing biases in recommendation results and improving the accuracy of personalized recommendations. This approach retains the model's expressiveness while avoiding additional cost burdens, and performs well in complex data environments with high-dimensional features, particularly suitable for personalized recommendation tasks that require handling multi-dimensional user data.

\subsubsection{Conversational Recommendation Systems}

As big data provides rich user behavior and interaction data for recommendation systems \cite{gan2017data}, and LLMs have enhanced their ability to understand and generate complex natural language instructions and dialogues \cite{lai2024large,zheng2024large}, the challenges faced by recommendation systems go beyond simply providing personalized recommendation content. They also need to understand better the dynamic changes in user needs and the diversity of conversational intentions. Conversational recommendation systems (CRS) \cite{chen2024data} provide personalized recommendations by simulating human dialogue, thereby establishing real-time user interactions to adapt to their constantly changing preferences. The core of CRS is effectively capturing users' dynamic and changing preferences. The introduction of MoE provides a more flexible architecture for CRS, allowing the training of gated networks for different dialogue intentions or scenarios. These networks can adaptively coordinate the outputs of different expert networks based on user input, generating recommendations that better meet user needs. The application of MoE models in CRS is more extensive, as they can effectively process and analyze large amounts of user interaction data, particularly in handling big data and improving recommendation efficiency, making them an important component in modern recommendation systems. Recent research has leveraged the knowledge understanding and reasoning capabilities of LLMs, converting user behavior sequences in sequential recommendation systems into fine-tuning prompts, and using low-rank adaptation (LoRA) modules to optimize recommendations. However, uniformly applying a single LoRA across different user behaviors can lead to indistinct individual differences, causing negative transfers between different users. Thus, Kong et al. \cite{kong2024customizing} combined the LoRA model with the MoE framework, creating a diverse group of experts who adaptively adjust model parameters at the instance level. Each expert is responsible for handling user-specific preferences, and the gating function dynamically adjusts the experts' participation based on the representation of the user's dialogue behavior sequence. This approach flexibly adapts the recommendation strategy, effectively mitigates the negative transfer effect, and improves the accuracy of big data conversation recommendations.

Current conversational recommendation methods often rely on recently observed user behaviors and user-item interaction data to build recommendation models. However, they fail to effectively capture long-term preferences and various auxiliary user information, which limits their recommendation performance. Therefore, incorporating multi-dimensional data such as web searches, user history, and product features into a hybrid expert model has become an area for improvement, providing a more comprehensive user profile through multi-dimensional data analysis to deliver more contextually relevant recommendations. For example, the hybrid approach model called HySAR \cite{bauer2024hybrid} for "conversation-aware" personalized recommendation treats various types of side information as model features, combining them with efficient tabular data machine learning methods to provide appropriate item recommendations in ongoing user conversations. Evaluations of HySAR on multiple public e-commerce datasets have demonstrated the effectiveness of its novel model features. These features encompass a range of information, including users, their actions, items, and conversational context, thereby improving the efficiency and adaptability of conversational recommendations.

\subsubsection{Cross-domain Recommendation}

Previous recommendation system research has typically focused on personalized recommendations within a single domain, unable to effectively utilize the associative data between cross-domain user behaviors and product information to provide more effective recommendations. This limitation has increasingly turned cross-domain recommendation systems into a research hotspot. Cross-domain recommendation systems can leverage the associative information between users and products in different domains to effectively expand the recommendation scope and improve recommendation performance. However, as the data scale continues to expand, traditional recommendation models struggle to handle the domain generalization and domain transfer issues in cross-domain recommendations. From the perspective of cross-domain data, while transferring knowledge between different domains is important, the challenge lies in how to effectively integrate multisource heterogeneous information to adapt to the complexity and diversity of multi-domain data. As a powerful neural network architecture, the MoE model provides new ideas for solving cross-domain recommendation problems in the big data context. In cross-domain recommendation tasks, MoE can train expert networks to handle data from different domains separately, and the gating network can dynamically blend the experts to adjust the recommendation strategy, achieving personalized recommendations across domains. In recent years, the research focus of cross-domain recommendations has gradually shifted from simple cross-domain data integration to how to effectively handle the heterogeneity between domains and avoid negative transfer. Early matrix factorization and collaborative filtering models have gradually been replaced by deep learning methods, particularly those that combine techniques like meta-learning and reinforcement learning with hybrid expert models. The Hypernetwork-based MoE (HMOE) model \cite{qu2022hmoe} in 2022 has been used for domain generalization problems. This method is not reliant on domain labels and has better interpretability. HMOE effectively identifies heterogeneous patterns in data by using hyper-networks to generate expert weights and has achieved good performance on multiple datasets, improving the domain generalization capability of recommendation systems. The Meta-DMoE framework \cite{zhong2022meta} combines meta-learning with knowledge distillation to optimize the knowledge transfer process through a forced aggregator. This effectively transfers knowledge from different source domains to the target domain, ensuring the accuracy and effectiveness of cross-domain recommendations. The introduction of meta-learning strengthens the model's generalization capability, and the knowledge distillation technique extracts positive knowledge for recommendations in the target domain, effectively addressing the risk of negative transfer. In the latest applications of cross-domain recommendations, the cross-domain sequential recommendation (CDSR) model \cite{park2024pacer} was proposed. It utilizes information from multiple domains to enhance recommendation performance, further advancing the solution to the domain transfer problem. CDSR dynamically evaluates the degree of negative transfer from each domain and uses it as a weight factor to control the gradient flow in domains with significant negative transfer, thereby avoiding the performance degradation caused by negative transfer. In extensive experiments across multiple domains, this model outperforms single-domain sequential recommendation models in recommendation performance.

\subsection{Cross-disciplinary Applications and Integration of MoE}

In addition to the major application areas mentioned above, MoE has also facilitated knowledge transfer and collaboration between different fields, demonstrating its powerful big data processing capabilities in a wider range of areas such as healthcare, financial risk management, and intelligent manufacturing, with promising application prospects.

\subsubsection{Medical Diagnosis}

In the field of medical diagnosis, the challenge of big data lies in how to extract effective information from complex and multidimensional medical data for accurate diagnosis. MoE provides a powerful tool for personalized disease risk assessment and multi-task diagnosis by dynamically selecting appropriate expert networks through the gating network to handle different patients' symptoms, offering efficient and accurate diagnosis and prediction support for various diseases. In medical research, clustering multi-dimensional time series data is a common goal for identifying different disease progression trajectories. Although methods that only cluster single-dimensional time series features have certain limitations, using models like MoE to integrate multiple time series features can reveal co-existing temporal patterns and generate deeper biological insights \cite{lu2023joint}. The MOELoRA framework \cite{liu2024moe} addresses the multi-task requirements and fine-tuning issues of LLMs in medical applications, combining the multi-task learning advantages of MoE and the efficient parameter fine-tuning advantages of LoRA to achieve efficient parameter fine-tuning in multi-task medical applications. MoE has also shown great potential in the diagnosis of medical imaging data. For example, Chen et al. \cite{chen2022automo} proposed a unified framework model called AutoMO-Mixer, which successfully achieved improved robustness and safety in optical coherence tomography (OCT) data diagnosis through a multilayer perceptron mixer and entropy-based evidential reasoning. In recent years, MoE has been used more extensively and efficiently for a wide range of difficult diseases, including spinal cord injuries \cite{liu2024empt}, cardiovascular disease, Alzheimer's disease \cite{zhang2024definition}, and non-Mendelian complex diseases \cite{courbariaux2022sparse}. This demonstrates that MoE technology has unique advantages in handling multi-dimensional and complex big data tasks in the healthcare domain, whether in image analysis, disease risk assessment, or the analysis of complex genetic information. This not only provides effective diagnostic tools for medical professionals but also promotes progress in clinical decision-making and patient care.

\subsubsection{Financial Risk Management}

For complex financial risk management problems, MoE is widely applied to assign different types of risk factors to expert networks for modeling, and the gating network dynamically coordinates the outputs of the experts to achieve more accurate risk assessment. With the development of big data technology, the scale and complexity of financial data have increased significantly, and traditional risk management tools are struggling to cope with this. Financial institutions have gradually adopted the MoE architecture to improve the accuracy and efficiency of risk assessment, helping them effectively identify and manage various risks. For example, the quantitative investment field is highly dependent on accurate stock forecasting and profitable investment decision-making. Although deep learning has made significant progress in capturing stock market trading opportunities, its sensitivity to random seeds and network initialization limits its widespread application. To address this issue, the AlphaMix framework \cite{sun2022quantitative} redefines the quantitative investment task as a multi-task learning problem, using a two-stage hybrid expert model to optimize the trading strategy. This framework optimizes multiple trading experts through personalized risk-aware loss functions in the two stages and then uses a neural router to dynamically schedule the experts, simulating the decision-making process of successful trading companies. AlphaMix performed well in long-term financial big data, surpassing existing benchmarks. In addition, MoE has shown excellent risk-sensitive investment decision-making capabilities in reinforcement learning research in finance (FinRL). For example, PRUDEX-Compass \cite{sun2023prudex} systematically evaluates the real-world application performance of FinRL methods across six dimensions and 17 indicators. The AutoEIS framework \cite{xiao2024autoeis} combines a multi-domain-aware expert mixture structure to significantly improve the accuracy of default prediction in the financial field, effectively solving the problems of numerical feature encoding and high-order feature interaction modeling in financial default prediction. The conditional evolution prediction of the Volterra process is a major challenge in mathematical finance, with its infinite dimensionality and non-smoothness posing significant obstacles for traditional models. To address this, a two-step solution was proposed recently \cite{arabpour2024low}. First, it uses stable dimensionality reduction techniques to project it into a low-dimensional statistical manifold and then uses a MoE-based model to dynamically update the parameters and effectively encode its non-stationary dynamics. The gating mechanism in MoE ensures that each expert focuses on specific time points, achieving an efficient approximation of the conditional distribution. This solution has significantly improved the prediction accuracy of the Volterra process, overcoming the limitations of the curse of dimensionality under big data.

\subsubsection{Intelligent Manufacturing Industry}

In the industrial field of global intelligent manufacturing development, the application of big data is focused on complex process parameters and equipment status. MoE can train expert networks for different process parameters or equipment states, and the gating network can adaptively select the appropriate experts based on real-time monitoring data to provide efficient decision support for intelligent manufacturing. Current research focuses on the prediction of industrial production status. The first parallel integrated deep neural network model (HDNN) \cite{al2019multimodal}, which combines two types of deep learning models (LSTM and CNN), has been used to improve the accuracy of remaining useful life (RUL) prediction for critical infrastructure and mechanical equipment, thereby significantly outperforming other existing methods in complex prediction scenarios. For industrial equipment life prediction and health management (PHM), a dual-task network structure combining a bi-directional gated recurrent unit (BiGRU) and multi-gate MoE (MMoE) \cite{souza2022contextual} has been utilized to simultaneously perform equipment health status assessment and RUL prediction. This approach automatically learns the optimal trade-off between HS assessment and RUL prediction, reducing the need for manual weight adjustment and demonstrating superior performance and robustness compared to existing models. Zhang et al. \cite{zhang2022health} introduced a new data-driven model called cMoE, which integrates process knowledge to enhance the effectiveness of human-machine collaboration in process industries. It not only improves predictive performance but also enhances the model's interpretability. For the progress of quality prediction in multistage manufacturing systems, Wang et al. \cite{wang2023production} introduced a multi-scale CNN with a multi-layer multi-gate expert hybrid multi-task model (ML-MMoE), which combined with soft parameter sharing and multi-gate attention mechanism, realizes simultaneous prediction of multi-task quality in all stages of a multistage manufacturing system, improves the quality prediction accuracy of each task, and provides more efficient multi-task quality prediction for industrial production. The solution is more efficient for multi-task quality prediction in industrial production. Recently, a multi-task physics-informed multi-gate MoE model \cite{yuan2024predicting} integrates physical information and multi-task big data modeling to provide higher-accuracy energy consumption prediction for hybrid energy supply systems, thereby driving the sustainable development of the intelligent manufacturing industry.

\section{Advantages and Challenges of MoE} \label{sec:advantages}
\subsection{Advantages}

\textbf{Scalability}. The modular design of the MoE framework makes it easy to extend and adjust. We can use different expert models to optimize for specific tasks or domains, and easily adapt to dataset changes ranging from small-scale to massive-scale by increasing the number of expert models, adjusting the weight ratios of expert models, etc. This allows the construction of more flexible, diverse, and scalable LLMs to handle larger datasets or more complex tasks, with an impressive performance on big datasets \cite{chen2023sparse,pham2024competesmoe,saikai2023mixtures}. On this basis, the Sparse Mixer algorithm \cite{liu2023sparse} provides accurate gradient approximation, further enhancing the scalability of MoE models when handling many tasks. The Sparse Mixer algorithm effectively bridges the gap between backpropagation and sparse expert routing, providing scalable gradient approximations for these critical steps, and ensuring reliable gradient estimation during MoE training. Especially in deep networks, the sparse routing mechanism can effectively reduce unnecessary computational overhead, so that the model scale is no longer limited by a single computational resource. This scalability capability is particularly important in the field of language processing \cite{bendale2024sutra,tang2024smile,zhou2024moe}, such as in multi-language processing and cross-cultural language understanding tasks, where the number and type of experts can be dynamically adjusted according to different task requirements. The model can flexibly switch between different languages, improving the extensibility of multi-language while maintaining high performance and flexibility in addressing complex semantics and language differences. Compared to other deep learning models, MoE stands out in its ability to adapt to task complexity and the growth of data scale. This design reduces the bottlenecks of MoE in handling many tasks, better meeting the model scalability requirements of modern machine learning.

\textbf{Efficient resource utilization}. Traditional fully connected networks need to activate all neurons or layers in each inference process, leading to a large amount of unnecessary computational overhead. Compared to traditional fully connected networks, those MoE models have very high sparsity since only a few expert models are activated, with most models in an inactive state. This sparsity brings improved computational efficiency, as the MoE model only activates a few most relevant experts through the gating mechanism, with only specific expert models processing the current input, significantly reducing the computational load and memory requirements. This selective activation allows MoE to reduce inference time and cost while maintaining efficient computation, effectively saving computational resources and providing users with faster AI response speeds \cite{he2023merging,zadouri2023pushing}. During the deployment of MoE models, the application of dynamic gating and load-balancing mechanisms further optimizes memory usage, making the model more suitable for deployment in resource-constrained environments \cite{huang2023towards}. Expert caching technology also provides the ability to cache expert model results \cite{skliar2024mixture}, further optimizing resource consumption by reducing redundant computations. In recent years, model optimization has further improved resource utilization. The elastic MoE training and fusion communication strategy proposed by SE-MoE \cite{yu2024moesys} has effectively improved the scalable inference performance on a single node. The end-to-end training and inference solution of DeepSpeed-MoE \cite{rajbhandari2022deepspeed} has achieved up to 9 times cost savings and faster inference speeds. Pre-gated MoE \cite{hwang2024pre} implements the gating mechanism in the model's preprocessing stage, effectively alleviating the dynamic computational burden brought by sparse expert activation and reducing the GPU memory requirements. Furthermore, the latest MoE++ architecture \cite{jin2024moe++}, through optimized "zero-compute experts", allows each token to use a variable number of FFN experts, and even completely skip some MoE layers. Compared to dense models, MoE++ instruction tuning can improve the performance of large models by approximately 45\% while using only 1/3 of the computational power, reducing training time. Additionally, the effectiveness increases with the parameter scale. In many machine learning tasks, resource utilization efficiency has become one of the important indicators to evaluate model superiority. MoE's unique activation and caching mechanisms give it a clear advantage in this area, making it an attractive choice for resource-sensitive applications.

\textbf{Better generalization ability}. In complex task scenarios, each expert only handles a specific data subset or domain, allowing for more detailed feature capture and analysis of the input data, achieving more effective feature extraction and prediction. By leveraging the strengths of each expert, MoE can handle complex and changing task scenarios, improving the overall model's generalization ability. Therefore, MoE can effectively avoid inter-task interference in multi-task learning, as each expert can focus on a specific task, thereby improving the performance of different tasks \cite{gupta2022sparsely,zhou2024adaptive}. Especially when there are differences in cross-domain data, MoE can leverage its expert specialization mechanism to flexibly adapt to the characteristics of different data distributions, improving performance in complex scenarios, and significantly outperforming traditional deep learning models. The generalization ability of the MoE model offers significant advantages in multi-task learning and cross-domain applications, primarily due to the specialization capability of its expert modules. For example, when a model faces the multi-task challenge of image recognition and natural language processing, the MoE model can accurately allocate these two types of tasks to the corresponding expert networks in their respective domains, avoiding conflicts between the data features of different tasks. Therefore, each expert can be trained and predicted on its specific data subset, improving the overall model's performance and fit. In addition, the MoE gating network dynamically evaluates the input data features to ensure that only the most suitable experts participate in the inference process. Compared to single-model architectures, MoE's layered and dynamic allocation mechanism gives it advantages in diversity and adaptability, ensuring stability and accuracy in changing task environments. Therefore, MoE has unique advantages in handling multi-task, multi-domain scenarios, and demonstrates excellent generalization ability.

\textbf{Powerful tools for big data processing}. MoE has become a prominent technology in the deep learning field due to its unique advantages in handling large datasets. The MoE approach, in which each expert network focuses on processing a specific data subset, effectively utilizes sparse matrix computation and leverages the parallel capabilities of GPUs \cite{he2021fastmoe,kamahori2024fiddler,pan2024parm} to accelerate the computation process. Since each expert only handles a specific sub-task, this parallel computing mode can maximize the utilization of computing resources, reducing computation time and memory consumption without sacrificing performance. Furthermore, as MoE allows multiple expert networks to operate simultaneously, the model scale can be flexibly adjusted according to the requirements, shortening training time and optimizing inference performance. Therefore, MoE can achieve better training efficiency and inference speed on large datasets compared to traditional methods, making it easy to scale to large datasets and provide better results at lower computational costs. Additionally, MoE ensures that the model's performance and resource efficiency are not affected as the dataset grows, an important feature for big data processing tasks. For example, in the field of natural language processing, MoE can allocate different language or semantic tasks to appropriate experts through its sparse deployment strategy, maintaining consistently high performance on text data in different languages. The sparse activation mechanism not only improves computational efficiency but also reduces computational costs, making MoE a mainstream choice for LLMs. Furthermore, the gating network dynamically allocates computational resources for the expert networks, further reducing computational overhead while ensuring model performance. These characteristics make the MoE approach a powerful tool for deep learning in big data environments, ensuring that the model can continue to provide high-performance services as the dataset grows.

\subsection{Challenges}

\textbf{Load imbalance and expert utilization}. In the MoE model, the primary function of the gating network is to select the most suitable experts to process a task based on the input features. If the design and training of the gating network are improper, it may lead to the problem of load imbalance \cite{antoniakmixture,li2024locmoe,wang2024home,zhou2022mixture}, where some experts are frequently called upon due to their superior performance, while others remain idle for an extended period. This imbalance not only wastes computational resources but also leads to over-training or stagnation of certain experts, reducing the overall model performance. The load imbalance mainly stems from the bias in the expert selection by the gating network, which may be caused by the uneven distribution of the training data or the architectural limitations of the gating network \cite{krishnamurthy2023improving}. To address the load imbalance issue, the existing research has proposed various strategies. Auxiliary loss to encourage load balancing \cite{dai2024deepseekmoe,fedus2022switch,wei2024skywork} is a common method that introduces additional loss terms during the training process to guide the gating network to distribute the experts more evenly and mitigate the load imbalance problem. Building upon this, the Loss-Free Balancing approach \cite{wang2024auxiliary} can effectively control the expert load without introducing interference gradients, leading to better performance and load balancing. Dynamic weight adjustment can also effectively balance the load \cite{huang2024harder,kong2024customizing,zhu2024dynamic}, where clustering algorithms can be used to dynamically adjust the weights of individual expert networks based on the changes in the sample data. Alternatively, the gated network can dynamically adjust the load based on the performance of the metadata server through meta-learning \cite{wu2020mdlb}. These methods can automatically adjust the cluster centers and membership degrees according to the data distribution, thereby optimizing the weight allocation of the expert networks. However, these methods also introduce new challenges, such as potential instability in the model training process, which require careful evaluation and adjustment in practical applications. Although these methods have to some extent alleviated the load imbalance problem, further research is needed to optimize the utilization of experts in larger-scale and more complex task scenarios, especially in distributed computing environments.

\textbf{Stability and training difficulty of gating networks}. Although the gating network in MoE models enables the intelligent allocation of tasks, the unstable gradient updates during the training process often lead the gating network to be biased towards certain experts, limiting the model's performance \cite{fan2022m3vit}. Particularly when the number of expert nodes is large, the problems of gradient vanishing or exploding become more pronounced, affecting the convergence and stability of the training \cite{hazimeh2021dselect,liu2022sparsity}. This can result in the model failing to converge or even training failure \cite{wu2024lazarus}. Although SparseMixer \cite{liu2023sparse} has introduced an approximate computation for the gradient terms often overlooked in typical MoE training--thereby accelerating the training convergence and mitigating the gradient issues--the routing strategy based on the multi-head attention mechanism may be overly complex, increasing the training difficulty and posing the risk of overfitting. Currently, the main source of the training difficulty of the gating network lies in its complex nonlinear structure and the need to optimize the parameters of multiple experts simultaneously \cite{krishnamurthy2023improving,sarkar2024block}. To address the training difficulty of the gating network, an innovative pre-training method was proposed to simplify the training process of the gating network \cite{zhong2024lory}. The Lory framework achieves a fully differentiable MoE architecture through expert merging techniques, solving the differentiability and training efficiency challenges of LLMs in the MoE paradigm. This work not only advances the research frontier of LLMs but also lays a solid foundation for building more efficient and powerful AI systems. In the future, to further improve the training efficiency of the MoE models, research must address the unstable gradient phenomena during the training process. As the model scale continues to expand, more advanced algorithms and system optimizations will be required \cite{yu2024moesys}.

\textbf{Collaboration and conflict management among experts}. In the MoE model, collaboration and conflict management among multiple experts are the key factors affecting the model's performance. The cooperative and competitive relationships among the experts in the MoE model need to be carefully managed to avoid the impact of output conflicts on the final prediction. Each expert may provide different outputs when processing a specific task, and these outputs need to be fused in the final decision-making. However, as each expert model has its domain knowledge and preferences, their understanding and processing of the same problem may differ, leading to different outputs. Some expert models may tend to be more conservative in their predictions, while others may be more aggressive, resulting in inconsistencies in the overall decision-making. This can affect the model's accuracy and robustness \cite{jiang2023mixture,wehenkel2022robust}. The collaboration problem among experts is mainly reflected in how to effectively integrate the outputs of different experts \cite{du2023redesigning}. The widely used attention-based methods have high data requirements, and if the data is noisy or insufficient, it may easily lead to performance degradation \cite{lee2021few,xue2021kdexplainer}. Therefore, reasonable collaboration mechanisms are needed. Current research has proposed various collaborative optimization strategies. For example, the innovative Mowst model \cite{zeng2023mixture} utilizes a "mixture of weak and strong experts" to handle graph neural network (GNN) tasks, encouraging the specialization and effective fusion of each expert. Furthermore, the conflict management strategies focus on how to handle the divergence among experts. The dilemma of expert weight allocation is a key issue \cite{dharmarathne2023shrinking,krishnamurthy2023improving}. When different experts provide completely different predictions for the same input, the gating network needs to make a wise weight allocation to ensure the accuracy of the final prediction. However, due to the differences in the specialized domains and confidence levels of different expert models, the precision of weight allocation is difficult to guarantee, which may lead to prediction bias \cite{makkuva2019breaking,verma2023learning}. Additionally, it is necessary to consider how to set a reasonable conflict detection mechanism. Some studies have introduced conflict detection mechanisms, where more complex decision-making processes \cite{chen2024llava,yang2024solving}, such as the participation of additional experts or the use of more complex fusion methods, are automatically triggered when significant conflicts are detected between experts. Although these methods have to some extent improved the model's stability and accuracy, further verification and improvement are still needed in practical applications.

\textbf{High requirement for data quality}. An important challenge faced in applying MoE models is the high requirement for data quality, as they exhibit a certain fragility in handling data noise. MoE models divide the data among different experts for processing, relying on the specialized capabilities of each expert for specific data subsets. If the input data contains significant noise, this structural advantage may be weakened, making it difficult for the model to correctly identify the optimal experts \cite{abbasi2016regularized,jung2018mixtures,yi2024variational}. Additionally, the weight allocation generated by the MoE model is also affected by data noise \cite{jiang2023mixture}. In the presence of significant noise, the model may mistakenly amplify the importance of certain irrelevant experts, leading to output instability. This not only reduces the model's interpretability but may also lead to a decrease in the model's generalization ability when facing unseen data. Compared to MoE models, some simple models are less sensitive to noise, allowing them to better capture the overall trends in big data and reduce the risk of overfitting. Therefore, in cases of lower data quality, choosing a MoE model may not be the optimal strategy.

\textbf{Deployment challenges and communication overhead}. The design of MoE models, where data is input to different experts for parallel processing, faces several challenges in actual deployment. The sparsity requirement of MoE models demands hardware that can support efficient parallel computing, which imposes higher requirements on hardware resources \cite{lepikhin2020gshard}. In many cases, to deploy MoE models in resource-constrained downstream tasks, the model needs to be pruned or simplified to reduce its parameter size and computational requirements \cite{chen2022task}, which can affect the model's performance and effectiveness. Additionally, this sparse structure incurs additional communication overhead \cite{fedus2022switch,imani2024mixture}, especially in environments with limited network bandwidth or uneven distribution of computing nodes, as parallel computation requires sacrificing a significant amount of computing efficiency. This high communication overhead not only affects real-time processing capabilities but may also prolong task execution times, further impacting the production process.

\section{Future Development Trends of MoE}  \label{sec:future}

With the continuous advancement of artificial intelligence, the MoE model, as a highly flexible and scalable paradigm, will undoubtedly have a broader development prospect in the future. With in-depth research and technological breakthroughs, the MoE model will face important development trends in the following aspects.

\subsection{Improvement of Model Generalization Capability}

One of the core challenges of the MoE model is how to maintain a high level of generalization capability when facing different tasks and data distributions. In the future, to further improve the generalization performance of MoE, we can explore new types of dynamic gating mechanisms. Currently, the typical MoE model mainly uses attention-based architectures as expert models, but this structure may have shortcomings when dealing with complex data. We can explore more diversified dynamic gating strategies, such as the MoE models based on multi-head attention mechanisms \cite{zhao2024mlp}, to help the model learn the complex relationships between different parts of the input sequence and adapt to the complex and changing big data environment. In addition, the existing MoE models can often only handle a single task or perform well in a specific field. To improve the generalization ability of the MoE model, we can optimize the transfer learning mechanism for cross-task and cross-domain scenarios, so that MoE can be flexibly applied in different application scenarios. For example, by pre-training large MoE models and transferring them to specific tasks, the adaptability and generalization performance of the model can be significantly improved. The CSII model \cite{shu2024adaptive} uses a mixture of dominant experts to serve all scenarios, effectively solving the problems of data imbalance and negative transfer, and greatly improving the cross-domain capability of the recommendation system. Similarly, the algorithm optimization of combined experts mitigates the catastrophic forgetting problem in multi-domain task incremental learning and can also improve the transfer learning capability \cite{park2024learning}. In the future, we can explore innovative transfer learning mechanisms to more effectively reduce the demand for training data on new tasks and accelerate the deployment and application of models in different application scenarios.

\subsection{Enhancement of Algorithm Interpretability}

In recent years, artificial intelligence has achieved many practical applications, and the demand for model interpretability has been increasing. The internal structure and dynamic expert selection mechanism of the MoE model are often very complex, making its decision-making process difficult to understand. Therefore, we can further explore the interpretability enhancement mechanism of MoE to improve the interpretability of its internal decision-making process. This will cover the following aspects:

\textbf{1) Optimization and interpretation of hierarchical structure}: The hierarchical structure of the MoE model allows each layer of expert models to play different roles in decision-making. In the future, we can optimize the hierarchical structure of MoE and introduce a multi-layer gating mechanism \cite{quan2024dmoerm} to enhance the clarity and interpretability of the decision-making process of each layer. By understanding the activation status of the experts in each layer, we can help users better understand how the model selects and combines expert models under different data inputs.

\textbf{2) Interpretability enhancement based on causal inference}: Causal inference can help us understand the decision-making mechanism of the MoE model, such as how the model's output is affected when a certain feature of the input data changes slightly \cite{jiang2024m4oe}. Applying causal inference techniques to the expert selection and gating strategies of the MoE model can reveal the behavioral logic of the model under different decision-making paths, providing users with more intuitive and understandable explanations.

\textbf{3) Combination of knowledge distillation and interpretability}: Knowledge distillation \cite{phuong2019towards} is a technique that transfers the knowledge of a complex model to a simple model. Through knowledge distillation, the decision-making process of the MoE model can be transferred to a simpler interpretable model (such as a decision tree), thereby enhancing its interpretability while ensuring model performance. This model combination can allow users to enjoy the high performance of the MoE model while understanding its decision-making logic through a simple model.

\subsection{Improvement of System Automation Level}

In practical applications, the deployment and maintenance costs of the MoE model are relatively high, mainly reflected in the need for continuous optimization and adjustment to maintain the model's performance. In the future, with the development of automation technology, the architecture design of the MoE system should tend to have a higher degree of intelligence and automation. Current research has already explored many strategies to improve the level of system automation, such as the automatic control technology based on expert networks (the L-MVAE model \cite{ye2021lifelong} that automatically adjusts the weights of the expert network and uses the most relevant experts to optimize the computational cost during the reasoning process), the self-adaptive gating mechanism based on the gating mechanism \cite{guo2024dynamic} and the self-supervised learning \cite{ruslim2023mixture}, as well as efficient training methods for automatic optimization of MoE \cite{jiang2024lancet,pan2024parm}. These flexible self-management mechanisms have effectively improved the deployment efficiency and user experience of the MoE model. In the future, we can continue to explore more novel automation methods, enhance the adaptive capability to changes in data distribution, delve deeper into the mechanisms that can self-detect and adapt to changes in data distribution, and improve the robustness of the model. In addition, we can also focus on developing more efficient training algorithms and hardware acceleration technologies to optimize computational efficiency.

\subsection{Privacy Protection Based on Big Data}

In the era of big data with increasing emphasis on data privacy, MoE provides new application solutions for privacy protection. Traditional centralized machine learning models need to centralize the dataset on a central server for training, which poses the risk of data leakage. The MoE models can divide the model into multiple expert modules and train them separately on different devices or institutions so that each device or institution only needs to train its expert model without sharing the original data with other institutions. This can effectively protect privacy while realizing large-scale and distributed machine learning, providing secure and reliable big-data analysis solutions for various industries. In recent years, the importance of data privacy issues in machine learning applications has been increasing. Several studies have made great progress in exploring the application technologies that combine MoE with various privacy protection technologies, such as federated learning \cite{chen2022federated,wu2023fedms} and differential privacy \cite{zhang2023fedbrain}. These methods have made significant breakthroughs in generalization and robustness on heterogeneous and non-standardized datasets, thus realizing efficient and secure distributed learning. However, whether these methods can achieve effective model learning and training while protecting privacy is still a problem worth studying. For example, secure multi-party computation \cite{jiang2021review} and edge computing \cite{shi2016edge}, which are critical technologies in big data privacy protection, have yet to be combined with MoE in this field. Therefore, in the future, MoE will have the opportunity to deeply integrate with more diverse and more comprehensive technologies in the field of data privacy protection, and construct distributed learning architectures. Under the trend of increasing emphasis on big data privacy, this method of training each expert model on different devices and sharing and integrating knowledge without sharing sensitive data will have significant importance in privacy-sensitive areas such as healthcare, financial risk control, and smart homes.

Here, we specifically discuss the application scenarios and specific application cases of MoE in the field of big data privacy protection:

\textbf{1) MoE framework based on federated learning}. Using federated learning \cite{chen2022federated}, the MoE framework can perform distributed machine learning while protecting data privacy \cite{Zec2020FederatedLU}. Each participant (such as different hospitals or financial institutions) can train their own expert models without the need to share sensitive raw data. The central gating network is responsible for coordinating these expert models and dynamically routing the input data to the appropriate experts to obtain the final prediction result. This federated MoE architecture can effectively utilize the distributed data resources while maximizing user privacy protection.

\textbf{2) MoE framework based on differential privacy protection}. In the future, we can propose a differential privacy-based MoE model for privacy-preserving big data analysis \cite{chen2023privacy}. During the model training process, by adding appropriate noise to the parameters of the experts and gating network, differential privacy protection can be achieved within a certain privacy budget. Therefore, this model not only protects personal privacy but also resists various privacy attacks while maintaining model performance.

\textbf{3) MoE framework based on secure multi-party computation}. Secure multi-party computation (SMC) \cite{jiang2021review} technology can be combined with the MoE framework to realize privacy-preserving big data analysis. In the SMC-MoE architecture, each participant retains their expert models and trains the gating network through the SMC protocol without the need to directly share the original data. The key challenge is how to ensure that the data privacy of each participant is not leaked, while still learning valuable knowledge from the distributed data sources.

\textbf{4) Decentralized MoE framework in edge computing}. In the edge computing \cite{shi2016edge} environment, the MoE framework can be deployed on different edge devices, forming a decentralized learning architecture. Each edge device can train its expert models and collaborate with other devices to jointly learn a global gating network. The key technologies of this decentralized MoE architecture aim to improve computational efficiency and response speed, while maximizing the protection of user privacy and utilizing sensitive data without the need to centralize it to a central server.

\renewcommand{\arraystretch}{1.5}
\begin{table*}[!h]
\tiny
\centering
\caption{Representative open-source models in 2024, detailing the number of activated and total parameters, the number of experts in each model, the context length, the GitHub STAR, and the model's license.}
\label{openMoE}
\begin{tabular}{|c|c|c|c|c|c|c|c|c|} 
\hline
\multirow{2}{*}{\textbf{Name}} & \multirow{2}{*}{\textbf{Time}} & \multirow{2}{*}{\textbf{Affiliation}} & \textbf{Params.}          & \multirow{2}{*}{\textbf{Context length}} & \textbf{Expert count}    & \multirow{2}{*}{\textbf{STAR}} & \multirow{2}{*}{\textbf{License}}                                                                       & \multirow{2}{*}{\textbf{Link}}                             \\ 
\cline{4-4}\cline{6-6}
                      &                       &                              & Total Activated~ &                                 & Total/Activated &                       &                                                                                                &                                                   \\ 
\hline
Grok-1                & 2024.03               & xAI                          & 314B~ 86B        & 8192                            & 8/2             & 49.7k                 & Apache 2.0                                                                                     & \url{https://github.com/xai-org/grok-1}                \\ 
\hline
DBRX                  & 2024.03               & Databricks                   & 132B~ 36B        & 32k                             & 16/4            & 2.5k                  & Databricks open model license                                                                  & \url{https://github.com/databricks/dbrx}              \\ 
\hline
Mistral-8x22B         & 2024.04               & Mistral AI                   & 176B~ 44B        & 65k                             & 8/2             & 9.8k                  & Apache 2.0                                                                                     & \url{https://github.com/mistralai/mistral-inference}    \\ 
\hline
DeepSeek-V2           & 2024.05               & Deepseek\_Ai                 & 236B~ 21B        & 128k                            & 160/6           & 3.8k                  & \begin{tabular}[c]{@{}c@{}}Code license: MIT\\Model license: model agreement\end{tabular}       & \url{https://github.com/deepseek-ai/DeepSeek-V2}      \\ 
\hline
Yuan 2.0-M32          & 2024.05               & IEIT                         & 40B~ 3.7B        & 16K                             & 32/2            & 182                   & \begin{tabular}[c]{@{}c@{}}Code license Apache 2.0\\Model license Yuan2.0 License\end{tabular} & \url{https://github.com/IEIT-Yuan/Yuan2.0-M32}          \\ 
\hline
Qwen2-57B-A14B        & 2024.06               & Alibaba                      & 57B~ 14B         & 64K                             & 72/16           & 10.9k                 & Apache 2.0                                                                                     & \url{https://github.com/QwenLM/Qwen2.5}                 \\ 
\hline
Skywork-MoE           & 2024.06               & Kunlun                       & 146B~ 22B        & -                               & 16/2            & 126                   & Skywork community license                                                                      & \url{https://github.com/SkyworkAI/Skywork-MoE }         \\ 
\hline
XVERSE-MoE-A36B       & 2024.09               & XVERSE                       & 255B~ 36B        & 8k                              & 64/2            & 37k                   & Apache 2.0                                                                                     &\url{https://github.com/xverse-ai/XVERSE-MoE-A36B}       \\ 
\hline
Hunyuan-Large         & 2024.11               & Tencent                      & 389B~ 52B        & 256k                            & 16/2            & 1.3k                  & \begin{tabular}[c]{@{}c@{}}Tencent Hunyuan \\ licensing agreement  \end{tabular}          & \url{https://github.com/Tencent/Tencent-Hunyuan-Large}  \\
\hline
\end{tabular}
\end{table*}

\subsection{Deep Integration of MoE and Other AI Technologies}

In the future, the deep integration of MoE models and other AI technologies will become one of the important trends in the development of artificial intelligence. This cross-domain innovative integration is expected to drive continuous breakthroughs of MoE in big data modeling and bring more technical innovations and development opportunities to various industries. The MoE models are not limited to a single AI technology; instead, they will be combined with other advanced AI technologies to form more comprehensive and powerful solutions, bringing technological breakthroughs to various industries. There have been some cases of the integration of MoE models and AI technologies:

\textbf{1) Generative AI} \cite{wu2023ai}: As an emerging technology, it has been widely used for different applications \cite{xing2024artificial}. Those frameworks enhanced by MoE can help to overcome the limitations of applications in physical layer communication security and improve communication security \cite{zhao2024enhancing}.

\textbf{2) Meta-learning}: It can be used to optimize MoE model parameters and enhance the model's rapid adaptation to new tasks. Combined with meta-learning, those MoE models can achieve accurate modeling of diverse data subspaces, and enhance their robustness when facing changes in data distribution.

\textbf{3) Reinforcement learning (RL)}: The combination of MoE and RL technology has been applied in some complex scenarios and has shown good results. For example, the optimized application of the combination of deep RL and MoE in drone system communication ensures physical layer security \cite{wong2024addressing}. Further application research based on multi-task RL has further improved overall performance \cite{cheng2023multi}. In the scenario of emotional support dialogue, the supporter model based on expert-mixed RL effectively realized the activation of positive emotions and maintained the coherence of the dialogue \cite{zhou2023facilitating}.

\subsection{Open-source Platforms and Ecosystem}

In recent years, the rapid development of open-source technology has had a far-reaching impact on innovation in many fields. In the future, it is expected that more and more open-source MoE frameworks and platforms will emerge. Open source not only helps the rapid iteration and popularization of technology, but also promotes the construction of the entire ecosystem, forming a virtuous interactive environment from developers to users. The research of MoE models cannot be separated from the support and promotion of open-source construction. Currently, although there are some open-source MoE frameworks, as listed in Table \ref{openMoE}, they are generally not mature enough, and the existing open-source MoE frameworks still have room for improvement in scalability, training efficiency, interpretability, evaluation methods, and application scenario support. While the open-source framework is promoting the rapid iteration of MoE technology, the construction of a more active open-source community ecosystem is also gradually improving. The open-source community has launched many excellent open-source MoE models and pre-trained models. We believe that more comprehensive and easy-to-use MoE open-source frameworks and learning platforms will emerge in the future. In addition, the open-source community will collaborate more deeply with leading companies in various industries to jointly explore and promote the application of MoE models in practical scenarios. Through the collaborative development of specific application demonstrations and solutions, the process of MoE models transitioning from the laboratory to the market can be accelerated, and their true value in commercial applications can be realized.

\section{Conclusion} \label{sec:conclusion}

In this review, we have conducted a comprehensive and in-depth analysis and summary of the latest literature on the MoE model in the context of big data. From the perspectives of the basic principles of MoE, algorithmic models, key technical challenges, and practical applications, we have conducted an in-depth review and analysis of the latest progress in this field. We have elaborated on the basic principles of MoE, analyzed its core ideas, and discussed the key components of the MoE architecture in detail based on the core ideas. We then conducted an in-depth analysis of the MoE architecture’s performance in addressing big data challenges, including key technical issues such as processing high-dimensional sparse data, integrating heterogeneous multisource data, achieving online learning, and providing interpretability. In addition, we have reviewed typical case studies of MoE in fields such as natural language processing, computer vision, and recommendation systems, demonstrating the wide applicability and powerful capabilities of MoE in these areas. Through the systematic summarization and analysis in this paper, we hope to comprehensively elaborate on the advantages and innovations of MoE in big data processing and provide theoretical and practical references for further promoting the application of MoE in practical applications. In the future, with the improvement in model generalization capability, enhanced algorithm interpretability, and higher system automation levels, MoE will play an increasingly important role in various fields, leading the processing and analysis of big data to new heights.

\bibliographystyle{cas-model2-names}

\bibliography{main.bib}

\end{sloppypar}
\end{document}